\documentclass[10pt,twocolumn,letterpaper]{article}

\usepackage{iccv}
\usepackage{times}
\usepackage{epsfig}
\usepackage{graphicx}
\usepackage{amsmath}
\usepackage{amssymb}
\usepackage{multirow}
\usepackage{booktabs}
\usepackage{subcaption}
\usepackage{xcolor, colortbl}
\usepackage{rotating}

\colorlet{dgreen}{green!50!black}
\colorlet{dred}{red!50!black}
\colorlet{dblue}{blue!70!black}

\usepackage{pifont}
\newcommand{\cmark}{\ding{51}}%
\newcommand{\xmark}{\ding{55}}%

\usepackage[pagebackref=true,breaklinks=true,letterpaper=true,colorlinks,bookmarks=false]{hyperref}

\iccvfinalcopy % *** Uncomment this line for the final submission

\def\iccvPaperID{12074} % *** Enter the ICCV Paper ID here

\ificcvfinal\fi

\begin{document}

%%%%%%%%% TITLE
\title{Best Practices in Active Learning for Semantic Segmentation}

\author{Sudhanshu Mittal* \\
University of Freiburg\\
Freiburg, Germany\\
{\tt\small mittal@cs.uni-freiburg.de}
% For a paper whose authors are all at the same institution,
% omit the following lines up until the closing ``}''.
% Additional authors and addresses can be added with ``\and'',
% just like the second author.
% To save space, use either the email address or home page, not both
\and
Joshua Niemeijer*\\
German Aerospace Center (DLR)\\
Braunschweig, Germany\\
{\tt\small Joshua.Niemeijer@dlr.de}
\and
J\"{o}rg P. Sch\"{a}fer\\
German Aerospace Center (DLR)\\
Berlin, Germany\\
{\tt\small Joerg.Schaefer@dlr.de}
\and
Thomas Brox\\
University of Freiburg\\
Freiburg, Germany\\
{\tt\small brox@cs.uni-freiburg.de}
}

\maketitle
% Remove page # from the first page of camera-ready.
\ificcvfinal\thispagestyle{empty}\fi

\def\thefootnote{*}\footnotetext{These authors contributed equally to this work}

%%%%%%%%% ABSTRACT
\begin{abstract}
%Purpose: 
Active learning is particularly of interest for semantic segmentation, where annotations are costly. Previous academic studies focused on datasets that are already very diverse and where the model is trained in a supervised manner with a large annotation budget.
% current issue with some examples
In contrast, data collected in many driving scenarios is highly redundant, and most medical applications are subject to very constrained annotation budgets.
% about our work
This work investigates the various types of existing active learning methods for semantic segmentation under diverse conditions across three dimensions - data distribution w.r.t. different redundancy levels,  integration of semi-supervised learning, and different labeling budgets.
% finding and contribution
We find that these three underlying factors are decisive for the selection of the best active learning approach. As an outcome of our study, we provide a comprehensive usage guide to obtain the best performance for each case. We also propose an exemplary evaluation task for driving scenarios, where data has high redundancy, to showcase the practical implications of our research findings.

%Active learning automatically selects samples for annotation from a data pool to achieve maximum performance with minimum annotation cost. 
%This is particularly critical for semantic segmentation, where annotations are costly.
%In this work, we show in the context of semantic segmentation that the data distribution is decisive for the performance of the various active learning objectives proposed in the literature. 
%Particularly, redundancy in the data, as it appears in most driving scenarios and video datasets, plays a large role.
%We demonstrate that the integration of semi-supervised learning with active learning can improve performance when the two objectives are aligned.
%Our experimental study shows that current active learning benchmarks for segmentation in driving scenarios are not realistic since they operate on data that is already curated for maximum diversity.
%Accordingly, we propose a more realistic evaluation task for driving scenario in which the value of active learning becomes clearly visible, both by itself and in combination with semi-supervised learning.
\end{abstract}

%%%%%%%%% BODY TEXT

% \input{1_introduction}
\section{Introduction}
\label{sec:intro}
\begin{table}[!t]
\centering
\begin{tabular}{|l|c|c|c|c|}
        \hline
       \multirow{2}{*}{Dataset$\downarrow$} & \multicolumn{4}{c|}{Annotation Budget}\\
        & \multicolumn{2}{c|}{Low} & \multicolumn{2}{c|}{High} \\ 
        \cline{2-5}
      Supervision $\rightarrow$ & AL  & SSL-AL & AL  & SSL-AL\\
      \hline
      Diverse      & \cellcolor{green!25}\cmark       & \cellcolor{green!25}\cmark  & \cellcolor{black!20}\cmark       &  \cellcolor{black!20}\cmark   \\
      Redundant    & \cellcolor{green!25}\cmark     & \cellcolor{green!25}\cmark  & \cellcolor{green!25}\cmark    & \cellcolor{green!25}\cmark   \\
      \hline
\end{tabular}
\caption{We study current active learning~(AL) methods for semantic segmentation over 3 dimensions - dataset distribution, annotation budget, and integration of semi-supervised learning~(SSL-AL). Green cells denote newly studied settings in this work. Previous AL works correspond to the grey cells. This work provides a guide to use AL under all the above conditions.}
\label{tab:teaser}
\end{table}
%Objective: 
The objective of active learning is the reduction of annotation cost by selecting those samples for annotation, which are expected to yield the largest increase in the model's performance.
%General motivation of AL for segmentation: 
It assumes that raw data can be collected in abundance for most large-scale data applications, such as autonomous driving, but annotation limits the use of this data.  Semantic segmentation is particularly costly, as it requires pixel-level annotations. Active learning is, besides weakly supervised and semi-supervised learning, among the best-known ways to deal with this situation. 

%Deep AL cycle: 
In a typical deep active learning process, a batch of samples is acquired from a large unlabeled pool for annotation using an acquisition function and is added to the training scheme. This sampling is done over multiple cycles until an acceptable performance is reached or the annotation budget is exhausted. 
%Types of AL methods: 
The acquisition function can be either a single-sample-based acquisition function, where a score is given to each sample individually or a batch-based acquisition function, where a cumulative score is given to the whole selected batch. Existing active learning methods for semantic segmentation assign the score to the sample either based on uncertainty~\cite{Golestaneh2020equal, Sinha2019VAdv,mackowiak2018cereals} or representational value~\cite{dual, Sinha2019VAdv, pmlr-v108-shui20a}.
%Current Benchmarks: 
Most AL methods in the literature are evaluated on datasets like PASCAL-VOC~\cite{Everingham2010PascalVOC}, Cityscapes~\cite{Cordts2016Cityscapes}, and CamVid~\cite{BrostowFC:PRL2008}. The shared attribute between these AL benchmark datasets is that they are highly diverse, as they were initially curated to provide comprehensive coverage of their corresponding domains. This curation process, however, is a sort of annotation because it is typically not feasible in an entirely automated way. 

%These AL benchmark datasets are notable for their high diversity, as they were initially curated to provide comprehensive coverage of their corresponding domains.

\begin{figure}[!t]
\includegraphics{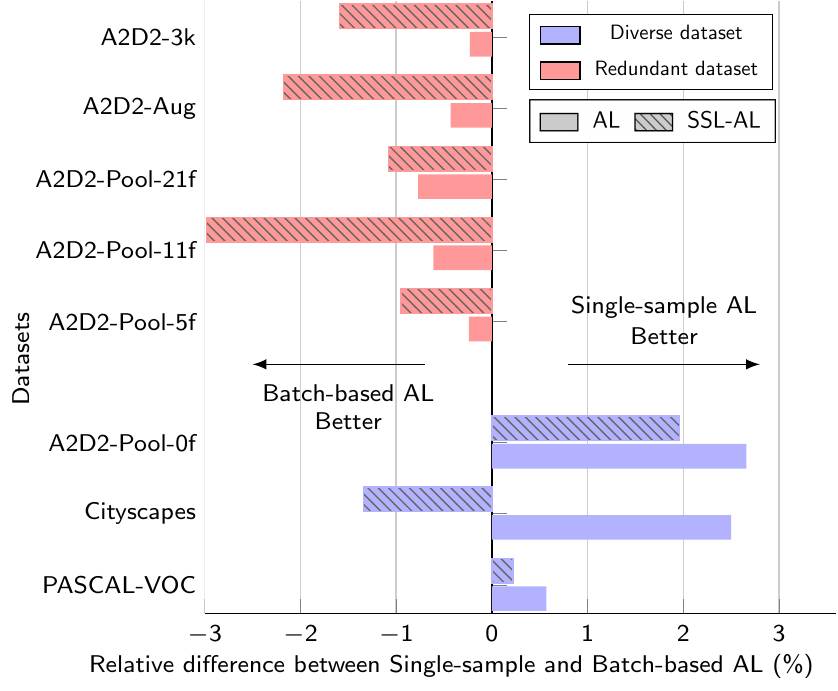}
\caption{We analyse and compare single-sample-based AL and batch-based AL on datasets with different levels of redundancy. The figure shows the difference between the best-performing single-sample-based AL method and best performing batch-based AL method. We find that batch acquisition performs better for redundant datasets, and single-sample acquisition performs better for diverse datasets. 
The integration of semi-supervised learning with active learning (SSL-AL) performs well for batch-based acquisition. 
}
\label{fig:teaser}
\end{figure}

%Point out the narrow scope of the evaluation: 
State-of-the-art active learning methods for segmentation have been evaluated only in a particular experimental setup - highly diverse benchmark datasets with a comparatively large annotation budget; see Table~\ref{tab:teaser}. 
%Raise Questions
We seek answers to specific missing questions not captured by previous works. \\
\textbf{1. How do different active learning methods perform when the dataset has many redundant samples?} Samples with highly overlapping information are referred to as redundant samples, for example, the consecutive frames of a video. 
Many commonly used segmentation datasets were originally collected as videos for practical reasons, e.g., Cityscapes, CamVid, BDD100k~\cite{YuBDD2018}. Since active learning methods were only tested on filtered versions of these datasets, their applicability on redundant datasets is open and highly relevant.\\
\textbf{2. What happens when the initial unlabeled pool is also used for training along with annotated samples using semi-supervised learning (SSL)?} For image classification, many works~\cite{mittal2019semi, consSSL, Huang_2021_ICCV, Munjal_2022_CVPR} have shown that integration of SSL into AL is advantageous. For semantic segmentation, this combination is not well studied.\\ 
%We notice that this integration for segmentation is more complex than seen for image classification due to the complexity of the task. \\
\textbf{3. What happens when the annotation budget is low? Which methods scale best in such low-budget settings?} Semantic segmentation annotations can be expensive for specific applications, especially in the medical domain. Therefore, it is critical to understand the behavior of the various active learning methods in low-budget settings.

%\textbf{Contribution.}
In this work, we report the results of an empirical study designed to find answers to the above-raised questions. 
%We provide a new way of approaching the active learning problem by questioning our knowledge about the given data and the annotation budget.
We study 5 existing active learning methods across the three dimensions as mentioned above - subject to different data distributions w.r.t. redundancy in the dataset, including the integration of semi-supervised learning, and under low as well as large annotation budget settings, as shown in Table~\ref{tab:teaser}. 
The outcome of this study yields new insights and provides, as the major contribution of this work, a guideline for the best selection of available techniques under the various tested conditions. Figure~\ref{fig:teaser} illustrates some of the results, particularly, that the performance of acquisition functions can change depending on whether the dataset is redundant or diverse and that SSL integration plays an additional role in this. Additionally, we show that active learning in a low annotation budget setting can be particularly volatile, even nullifying the complete need for it in some cases. This further emphasizes the importance of knowing the underlying data distribution.% can play a major role in method selection.

We also suggest a new evaluation task (A2D2-3K) for driving scenarios based on the highly redundant A2D2 dataset, which is closer to the raw data collection scheme in a driving case. The experiment outcome on this task aligns with the findings of our study for redundant dataset type with a high annotation budget setting and shows that there is a strong case for using active learning in this context.

\section{Deep Active Learning}\label{sec:sotawork}

In this section, we briefly review the state of the art in deep active learning as relevant for our study. In particular, we review the available acquisition methods, the special considerations for segmentation, and the integration of semi-supervised learning. 

The acquisition methods can be categorized into single-sample-based and batch-based approaches. They assess the value of new samples for selecting individually and collectively as a batch, respectively. %Various ways were suggested in previous works, such as model uncertainty, data coverage, and mutual information.

\textbf{Single sample acquisition} takes the top $b$ samples according to the score of the acquisition function to select a batch of size $b$. Several methods follow this selection scheme based on either epistemic uncertainty or representation score. 
For example, uncertainty-based methods try to select the most uncertain samples to acquire a batch. Many methods, such as EqualAL~\cite{Golestaneh2020equal}, Ensemble+AT~\cite{NIPS2017_7219}, and CEAL~\cite{Wang_2017_CAL_3203306_3203314}, estimate uncertainty based on the output probabilities. Epistemic uncertainty, estimated using Entropy~\cite{shannon_2001}, is often used a as strong baseline in several active learning works~\cite{Golestaneh2020equal, Shin_2021_ICCV, Rangnekar_2023_WACV}.
Some methods, namely BALD~\cite{Houlsby2011BALD} and DBAL~\cite{Gal:2017:DBA:3305381.3305504} employed a Bayesian approach using Monte Carlo Dropout~\cite{MonteDrop} to measure the epistemic uncertainty.
Representation-based methods aim to select the most representative samples of the dataset that are not yet covered by the labeled samples. Numerous adversarial learning-based methods utilize an auxiliary network to score samples based on this measure, including DAAL~\cite{dual}, VAAL~\cite{Sinha2019VAdv}, and WAAL~\cite{pmlr-v108-shui20a}. For our study, we employ Entropy, EqualAL, and BALD to represent single-sample acquisition methods due to their direct applicability to segmentation tasks. We did not include deep ensemble-based methods due to their limited scalability and adversarial methods due to their hyperparameter sensitivity. In general, single-sample acquisition approaches select individually very informative samples but do not optimize the joint improvement obtained with the whole batch.
 %Although, they can also be good candidates for comparison.
%\tb{Which ones do you select for the study and why?}

\textbf{Batch-based acquisition} methods acquire the whole batch of size $b$ to maximize cumulative information gain. 
Sener~\etal~\cite{sener2017active} formulated the acquisition function as a core-set selection approach based on the feature representations. It is a representation-based approach that selects the batch of samples jointly to represent the whole data distribution.
BatchBALD~\cite{Kirsch2019BatchBALD} is a greedy algorithm that selects a batch of points by estimating the joint mutual information between the whole batch and the model parameters. This method was also proposed to remedy the mode collapse issue, where the acquisition function collapses into selecting only similar samples (see Section~\ref{sec:res:data} for details).
%which is discussed in Section~\ref{sec:res:data}.
However, it is limited to simple image classification datasets like MNIST~\cite{deng2012mnist} since its computation complexity grows exponentially with the batch size. %\tb{Why is it limited to MNIST?}
Some more recent batch-based methods include k-MEANS++~\cite{DBLP:journals/corr/abs-1901-05954}, GLISTER~\cite{glister}, ADS~\cite{pmlr-v89-jia19a}, but these methods only evaluate on image classification tasks. For the study, we selected the Coreset method~\cite{sener2017active} to represent batch-based methods due to its effectiveness, simplicity, and easy scalability to the segmentation task. %\tb{Didn't you also work with BALD?}

\subsection{Active Learning for Semantic Segmentation}
When applied to semantic segmentation, active learning methods must choose which area of the image is to be considered for the acquisition: the full image~\cite{Sinha2019VAdv}, superpixels~\cite{Cai_2021_CVPR}, polygons~\cite{Mittal2019Illusions, Golestaneh2020equal}, or each pixel~\cite{Shin_2021_ICCV}. There is no common understanding so far of which approach is cheaper and more effective. Thus, our study uses the straightforward image-wise selection and annotation procedure. 

%\tb{Most of this looks redundant to what was said above already. Focus on the specifics of segmentation.}
%where the score is usually evaluated per pixel and averaged over all pixels in the image.
%\textbf{Entropy}~\cite{shannon_2001} (estimated uncertainty) is one such widely used information measure for selection. This function computes per-pixel entropy for the predicted output and uses the averaged entropy as the final score. This method selects the top-k high scoring images.
Most existing methods for segmentation are based on the model’s uncertainty for the input image, where the average score over all pixels in the image is used to select top-k images.
\textbf{Entropy}~\cite{shannon_2001} (estimated uncertainty) is a widely used active learning baseline for selection. This function computes per-pixel entropy for the predicted output and uses the averaged entropy as the final score.
%This acquisition function uses per-pixel entropy as an estimation of the epistemic uncertainty for the predicted output. The final score is the average entropy over all pixels. 
\textbf{EqualAL}~\cite{Golestaneh2020equal} determines the uncertainty based on the consistency of the prediction on the original image and its horizontally flipped version. The average value over all the pixels is used as the final score. 
%employs a self-consistency regularizer between equivariant views and uses the cumulative entropy of the views as the uncertainty measure.
\textbf{BALD \cite{Houlsby2011BALD}} is often used as baseline in existing works.
It is employed for segmentation by adding dropout layers in the decoder module of the segmentation model and then computing the pixel-wise mutual information using multiple forward passes. 
%\textbf{Coreset~\cite{sener2017active}} is a batch-based approach which was initially proposed for image classification, but it can be easily modified for segmentation. %It selects a batch of samples that cover the whole data distribution. 
%It formulates this batch selection as a robust k-center selection problem. 
%Coreset implements a greedy algorithm that iteratively selects unlabeled samples with the largest minimum distance to the labeled samples.
\textbf{Coreset~\cite{sener2017active}} is a batch-based approach that was initially proposed for image classification, but it can be easily modified for segmentation. For e.g., the pooled output of the ASPP~\cite{DBLP:journals/pami/ChenPKMY18} module in the DeepLabv3+~\cite{deeplabv3+} model can be used as the feature representation for computing distance between the samples.
%selects unlabeled samples with maximum distance to the nearest neighbour of the so far selected samples.
Some other methods~\cite{Sinha2019VAdv, Kim_2021_CVPR, pmlr-v108-shui20a} use a GAN model to learn a combined feature space for labeled and unlabeled images and utilize the discriminator output to select the least represented images. Our study includes Entropy, EqualAL, BALD, and Coreset approaches for the analysis, along with the random sampling baseline. In this work, these methods are also studied with the integration of semi-supervised learning. 
%Some hybrid approaches~\cite{dual} combine the representation score with uncertainty. 
%Most AL methods for semantic segmentation show that single-sample acquisition is superior to batch-based methods.

% \paragraph{Experiment baselines:}

% We use the EqualAL implementation, which trains using only cross-entropy loss to keep the baselines comparable.

% -------------------------------- Semi-supervised Active Learning ---------------------------------

\subsection{Semi-supervised Active Learning} \label{sec:ALforSemSeg:SemiSuper}
%SSL and AL are closely related tasks that aim to achieve the best model performance by using a combination of labeled and unlabeled samples.
Active learning uses a pool of unlabeled samples only for selecting new samples for annotation. However, this pool can also be used for semi-supervised learning (SSL), where the objective is to learn jointly from labeled and unlabeled samples.
The combination of SSL and AL has been used successfully in many contexts, such as speech understanding~\cite{karlos2021classification, Drugman2016}, image classification~\cite{sener2017active, consSSL, Mittal2019Illusions, Munjal_2022_CVPR}, and pedestrian detection~\cite{RHEE2017109}. 
Some recent works have also studied active learning methods with the integration of SSL for segmentation, but their scope is limited only to special cases like subsampled driving datasets~\cite{Rangnekar_2023_WACV} or low labeling budget~\cite{Mittal2019Illusions}, both cases with only single-sample acquisition methods.
Our work provides a broader overview of the integration of SSL and active learning for the segmentation task. We study this integration over datasets with different redundancy levels, under different labeling budgets, and with single-sample and batch-based methods. Our findings explain when this integration is effective and boosts the active learning method. 

%A successful integration can also be explained by the underlying assumption of SSL and the selection principle of the active learning methods, described below in detail. \\
%They are also conceptually supported by the underlying assumption of SSL and the selection principle of the active learning methods, described below in detail. 

%Active learning methods use a pool of unlabeled samples only for selecting new samples for annotation. However, this pool can also be used for semi-supervised learning, where the objective is to learn jointly from labeled and unlabeled samples. 
%In this work, we integrate semi-supervised learning with active learning in the context of semantic segmentation, an idea that was previously proposed for classification~\cite{sener2017active, consSSL, Mittal2019Illusions, Munjal_2022_CVPR}.

%\noindent \textbf{Integration of SSL and AL: Conceptual considerations.}
%\tb{Can this be integrated more tightly into the previous paragraph, such that it does not look like an addendum?} \sud{Done. Is it better now? }
%Here, we discuss some theoretical considerations that are vital for a successful integration.
\textbf{Integration of SSL and AL.} A successful integration can also be conceptually explained based on the underlying assumption of semi-supervised learning and the selection principle of the active learning approach.
According to the \textit{clustering assumption} of SSL, if two points belong to the same cluster, then their outputs are likely to be close and can be connected by a short curve~\cite{books/mit/06/CSZ2006}.
In this regard, when labeled samples align with the clusters of unlabeled samples, the cluster assumption of SSL is satisfied, resulting in a good performance. Consequently, to maximize semi-supervised learning performance, newly selected samples must cover the unlabeled clusters that are not already covered by labeled samples. Only acquisition functions that foster this coverage requirement have the potential to leverage the additional benefits that arise from the integration of semi-supervised learning.
A batch-based method, e.g., Coreset, selects samples for annotations to minimize the distance to the farthest neighbor. 
By transitivity, such labeled samples would have a higher tendency to propagate the knowledge to neighboring unlabeled samples in the cluster and utilize the knowledge of unlabeled samples using a semi-supervised learning objective and help boost the model performance.
%help in boosting the model performance. 
Similar behavior can also be attained using other clustering approaches that optimize for coverage. %However, k-means clustering is usually sub-optimal, particularly for small k, since it does not optimize coverage but average representation quality. Thus, rare samples are mostly ignored. \sud{paragraph ending missing?}
%\tb{Paragraphs should end with something conclusive or with a problem that is directly picked up in the next paragraph. Your paragraphs often end with a sentence where the reader thinks "and now what?", and then nothing follows.}

%\label{sec:sotawork}
% \input{3_method}
% \section{Methods}
% \input{chapters/methods}
%\section{Conceptual Considerations}
%\input{chapters/DeepLearning_AL}
%\input{chapters/methods}
%\label{sec:AL_in_DL}
% \input{4_experiments}
\section{Experimental Setup}

% \subsection{Baseline Methods}
\subsection{Tested Approaches}
In our study, we test five active learning acquisition functions as discussed in Section~\ref{sec:sotawork}, including Random, Entropy, EqualAL, BALD, and Coreset. Here Entropy, EqualAL, and BALD approach represent single-sample, and Coreset represents the batch-based approach. All methods select the whole image for annotation. To leverage the unlabeled samples, we use the semi-supervised learning s4GAN method~\cite{mittal2019semi}. 
%proposed by Mittal~\etal~\cite{mittal2019semi}. 
We pair all the used active learning approaches with SSL using this approach. This is marked by the suffix `-SSL' in the experiments. In particular, we train the model using an SSL objective, which impacts the resulting model and hence the acquisition function.

\subsection{Datasets}
Active learning methods are often evaluated on PASCAL-VOC and Cityscapes datasets, where PASCAL-VOC is naturally diverse while Cityscapes is diversified by subsampling from videos. In this work, we test on an additional driving dataset, A2D2, which is highly redundant. We evaluate the methods on these three datasets. 
To understand the nature of active learning methods over varying levels of redundancy in the dataset, we curate 5 smaller dataset pools from the large, original A2D2 dataset, described further below as A2D2-Pools.

%We evaluate AL methods on three semantic segmentation datasets: A2D2, Cityscapes, and PASCAL-VOC. To conduct our experiments, we curated 5 datasets from the large A2D2 data. Cityscapes and PASCAL-VOC are used in their original format. We train AL methods on the training sets and evaluate them on the corresponding validation sets. 
%We run all methods on 3 random seeds and report the mean performance. We evaluate AL methods across different data budget settings, denoted by I-S, where I is the initial label budget, S is the sampling-label budget, and I, S denotes the percentage of the dataset size. Images are sampled randomly to fulfill the initial label budget. For the subsequent steps, images are sampled using the AL acquisition function up to the allowed sampling-label budget. We test these datasets with 10-10, 5-5, and 2-2 budget settings. 

\textbf{Cityscapes}~\cite{Cordts2016Cityscapes} is a driving dataset used to benchmark semantic segmentation tasks. The dataset was originally collected as videos from 27 cities, where a diverse set of images were selected for annotation. Due to the selection, Cityscapes cannot cover the redundant data scenario in our evaluation, although it was derived from videos. As we will see in the results, the nature of the active learning method changes when considering the raw form of data in a driving scenario, and pre-filtering, as done in Cityscapes, is sub-optimal compared to directly applying active learning on the raw data (see Section~\ref{sec:res:a2d2_3k}).

%\textbf{Cityscapes}~\cite{Cordts2016Cityscapes} is a driving dataset used to benchmark semantic segmentation task. It consists of 3475 frames, where 2975 images are training images, and 500 images are used for validation. These images %have a resolution of $1024 \times 2048$ and 
%are labeled with 19 classes. The dataset contains images from 27 cities, where a diverse set of images were selected for annotation. \tb{I think you can drop details from the datasets (people know them or can look up the numbers) as long as you say what is special and relevant about the dataset for the study.}

\textbf{PASCAL-VOC}~\cite{Everingham2010PascalVOC} is another widely used segmentation dataset. We use the extended dataset~\cite{6126343}, which consists of 10582 training and 1449 validation images.
%The extended dataset consists of 10582 training images and 1449 validation images. 
%The dataset consists of 21 classes, including the background class. 
It contains a wide spectrum of natural images with mixed categories like vehicles, animals, furniture, etc. It is the most diverse dataset in this study.

\begin{figure}[]
\centering
\begin{tabular}{c@{\hspace{1mm}}c}
 \includegraphics[trim={0 0cm 0 0cm}, clip, width=70mm]{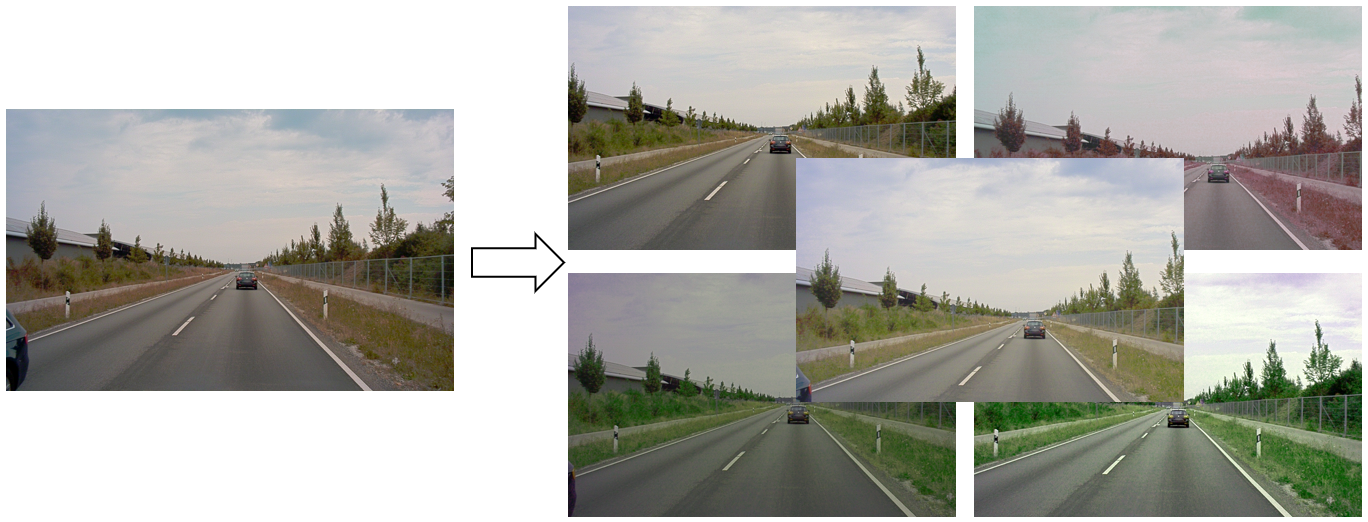}
\end{tabular}
\caption{A2D2 Pool-Aug. Left: the original image. Right: the duplication through color augmentation and random cropping of the original image}
\label{fig:Crop_and_augment}
\end{figure}

\textbf{A2D2}~\cite{geyer2020a2d2} is a large-scale driving dataset consisting of 41277 annotated images with a resolution of $1920\times1208$ from 23 sequences. It covers an urban setting from highways, country roads, and three cities. It contains labels for 38 categories. We map them to the 19 classes of Cityscapes for our experiments. A2D2 provides annotations for every $\sim10^{th}$ frame in the sequence and contains a lot of overlapping information between frames. Some consecutive frames are shown in the Appendix. We utilize 40135 frames from 22 sequences for creating our training sets and one sequence consisting of 1142 images for validation. The validation sequence `\textit{20180925\_112730}' is selected based on the maximum class balance. A2D2 represents the most diverse raw dataset in our study. 

 \begin{figure*}[t!]
\centering
\begin{tabular}{c@{\hspace{1mm}}c}
\begin{subfigure}{0.5\textwidth}
 \includegraphics[trim={0 0.5cm 0 0cm}, clip, width=75mm]{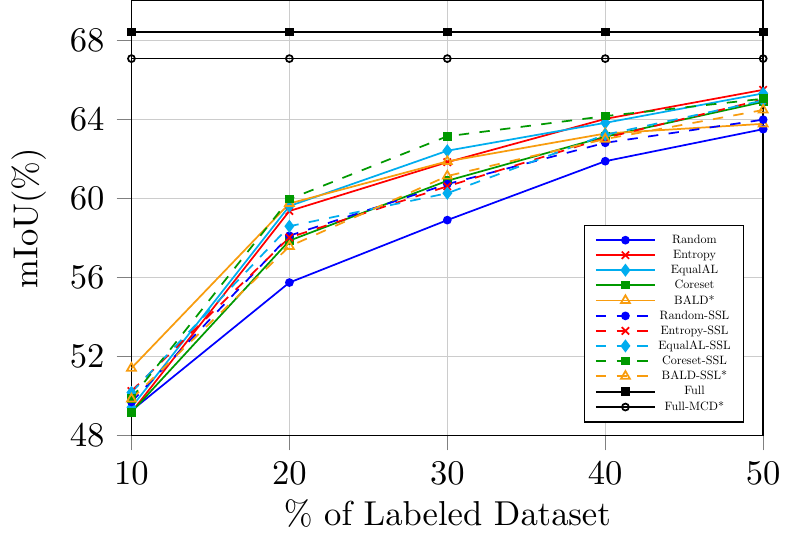}  \caption{Cityscapes}
 \end{subfigure} &
 \begin{subfigure}{0.5\textwidth}
  \includegraphics[trim={0 0.5cm 0 0cm}, clip, width=75mm]{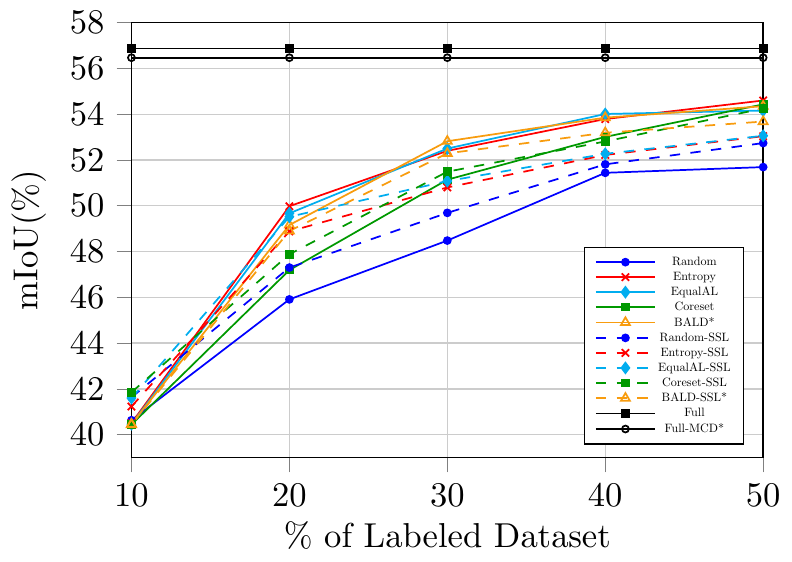} \caption{A2D2:Pool-0f}
 \end{subfigure} 
\end{tabular}
\caption{Results on diverse driving datasets. Active learning performance curves on Cityscapes and A2D2:Pool-0f. X-axis shows the percentage of the labeled dataset. The methods which utilize MC-dropout in their network architecture are marked with $\ast$, and are only comparable to other methods with MC-dropout. }
\label{fig:AL_diverse}
\end{figure*}

\textbf{A2D2 Pools.} To obtain a more continuous spectrum between diverse and redundant datasets, we created five smaller dataset pools by subsampling the large A2D2 datasets. Each pool comprises 2640 images, which is comparable in size to the Cityscapes training set.
Four pools are curated by subsampling the original dataset, while the fifth pool is created by augmentation. The first four pools, denoted by Pool-Xf (where X is 0, 5, 11, and 21), were created by randomly selecting samples and X consecutive frames for each randomly selected sample from the original A2D2 dataset. Pool-0f contains only randomly selected images. We assume that the consecutive frames contain highly redundant information. Therefore, the pool with more consecutive frames has higher redundancy and lower diversity. The fifth pool, Pool-Aug, contains augmented duplicates in place of the consecutive frames. We create five duplicates of each randomly selected frame by randomly cropping 85\% of the image area and adding color augmentation (see figure \ref{fig:Crop_and_augment}).\\

\begin{table}
\centering
\setlength\tabcolsep{3.0pt}
 \begin{tabular}{|c|l|@{\hspace{0.15cm}}c@{\hspace{0.15cm}}|@{\hspace{0.15cm}}c@{\hspace{0.15cm}}c@{\hspace{0.15cm}}|@{\hspace{0.15cm}}c@{\hspace{0.15cm}}c@{\hspace{0.15cm}}|} 
 \hline 
  \multirow{2}{*}{\textbf{A}} & \textbf{AL Method}   & \multirow{2}{*}{\textbf{SSL}} & \multicolumn{2}{c@{\hspace{0.15cm}}|@{\hspace{0.15cm}}}{\textbf{Cityscapes}}   & \multicolumn{2}{c@{\hspace{0.15cm}}|}{\textbf{A2D2 Pool-0f}}    \\
   &   \textbf{Metric} $\to$ &  & \textbf{mIoU\small}  & \textbf{AUC\small}  & \textbf{mIoU\small}  & \textbf{AUC\small} \\
 [0.5ex] 
 \hline
S & Random        & \xmark         & 58.90          & 23.29          & 48.48              & 19.20  \\ 
S & Entropy       & \xmark         & 61.83          & 24.25          & 52.40              & \textbf{20.37} \\
S & EqualAL       & \xmark         & 62.41          & 24.32          & \textbf{52.50}     & 20.35   \\
B & Coreset       & \xmark         & 60.89          & 23.89          & 51.14              & 19.88   \\
S & Random-SSL    & \cmark         & 60.72          & 23.85          & 49.69              & 19.60   \\
S & Entropy-SSL   & \cmark         & 60.61          & 23.93          & 50.80              & 19.90   \\
S & EqualAL-SSL   & \cmark           & 60.26          & 23.96          & 51.08              & 20.02   \\
B & Coreset-SSL   & \cmark           & \textbf{63.14} & \textbf{24.47} & 51.49              & 20.02   \\
- & 100\%         & \xmark           & 68.42          & 27.37          & 56.87              & 22.75   \\
\hline
\multicolumn{7}{|l|}{\textit{With MC-Dropout decoder} } \\
\hline
S & BALD          & \xmark         & 61.87          & 24.28             & \textbf{52.82}    & \textbf{20.32} \\
S & BALD-SSL    & \cmark       & 61.13          & 23.89             & 52.29             & 20.14     \\
%B & Coreset-MCD            & 60.60          & 23.78             & 49.99             & 19.45     \\
%B & Coreset-MCD-SSL        & \textbf{62.24} & \textbf{24.37}    & 51.76             & 19.97     \\
- & 100\%-MCD      & \xmark        & 67.07          & 26.83             & 56.47             & 22.59      \\
\hline 
\end{tabular}

\caption{Active Learning results on Cityscapes and A2D2 Pool-0f. AUC@50 and mIoU@30 metrics are reported. A denotes Acquisition method type. S and B denotes the single-sample and batch-based acquisition, respectively. }
\label{table:sota_results_cityscapes}
\end{table}

\begin{table}
\centering
\setlength\tabcolsep{3.5pt}
 \begin{tabular}{|@{\hspace{0.15cm}}c@{\hspace{0.15cm}}|@{\hspace{1.5mm}}l@{\hspace{1.5mm}}|@{\hspace{0.10cm}}c@{\hspace{0.10cm}}|@{\hspace{0.15cm}}c@{\hspace{0.15cm}}c@{\hspace{0.15cm}}|@{\hspace{0.15cm}}c@{\hspace{1.5mm}}c@{\hspace{0.15cm}}|} 
 \hline 
 \multirow{2}{*}{\textbf{A}} & \textbf{AL Method}   & \multirow{2}{*}{\textbf{SSL}} & \multicolumn{2}{c|@{\hspace{1.5mm}}}{\textbf{VOC 5-5}}   & \multicolumn{2}{c|}{\textbf{VOC 10-10}}   \\  
  
%    \cmidrule(l{0.0em}r{4.0em}){2-3}  \cmidrule(r{4.0em}){4-5} \cmidrule(r{0.5em}){6-7}
    & \textbf{Metric} $\to$  &   & \textbf{mIoU}  & \textbf{AUC }  & \textbf{mIoU}  & \textbf{AUC } \\
 [0.5ex] 
 \hline
S & Random      &  \xmark         & 70.70              & 13.92             & 72.13           & 28.85    \\ 
S & Entropy      &  \xmark        & 70.38              & 13.94             & 73.72           & 29.17    \\
S & EqualAL       &  \xmark        & 69.14              & 13.82             & 73.40           & 29.03   \\
B & Coreset        & \xmark        & 70.85              & 13.96             & 73.63           & 29.06     \\
S & Random-SSL      &  \cmark     & 72.57              & 14.36             & 75.33           & 29.87   \\
S & Entropy-SSL     &  \cmark     & 73.36              & 14.51             & \textbf{76.08}  & 30.01    \\
S & EqualAL-SSL     &  \cmark   & \textbf{73.39}       & \textbf{14.55}    & 75.89           & \textbf{30.06}      \\
B & Coreset-SSL     &  \cmark     & 72.88              & 14.46             & 75.91           & 30.03    \\
\hline
- & 100\%           & \xmark       & 77.00              & 15.40             & 77.00           & 30.80    \\ 
\hline
\end{tabular}

\caption{Active Learning results on PASCAL-VOC dataset in 5-5 and 10-10 settings. AUC@50 and mIoU@30 metric are reported. A, S and B denotes acquisition method type,  single-sample and batch-based acquisition, respectively.}
\label{table:sota_results_pasca_voc}
\end{table}

%\noindent \textbf{Diverse vs. Redundant datasets.}
%We categorize the above mentioned 7 datasets into diverse and redundant datasets for the analysis purposes. We consider PASCAL-VOC, Cityscapes, and A2D2:Pool-0f datasets as diverse datasets and A2D2:Pool5f, Pool-11f, Pool-21f, and Pool-Aug datasets as redundant datasets for our study. PASCAL-VOC can be tagged easily as diverse, whereas A2D2-Pool-5f, 11f, 21f, Aug can be easily tagged as redundant, respectively. In the middle of the spectrum, putting a clear redundancy/diversity tag is difficult. Cityscapes or A2D2-Pool-0f fall in this spectrum since they are curated by sparsely selecting from large video stream data. We consider them as non-redundant/diverse for our study as they behave analogous to diverse datasets.

%--------------------------------------------------

\subsection{Experiment Details}
\textbf{Implementation details.}
We used the DeepLabv3+~\cite{deeplabv3+} architecture with the Wide-ResNet38 (WRN-38) \cite{wide38} backbone for all experiments. The backbone WRN-38 is pre-trained using ImageNet~\cite{5206848}. We run all methods on 3 random seeds and report the mean performance. All other training details and hyperparameter information is included in the Appendix.
Since the BALD method requires the introduction of Dropout layers into the architecture, we segregate the methods into two categories: with Monte Carlo Dropout (MCD) and without Monte Carlo Dropout layers. Random, Entropy, EqualAL, and Coreset are without MCD. BALD is based on a MCD network. Since the models used in the two categories are not exactly comparable due to different architectures, 
%we show performance of BALD both with and without SSL integration for completeness. 
we also show the fully-supervised performance for both with MCD~(100\% MCD) and without MCD~(100\%) architectures.

\begin{table*}
\centering
 \begin{tabular}{|c|l@{\hskip 0.15in}|c|c@{\hskip 0.1in}c|c@{\hskip 0.1in}c|c@{\hskip 0.1in}c|c@{\hskip 0.1in}c|} 
 \hline 
  \textbf{A} & \textbf{AL Method}  & \multirow{2}{*}{\textbf{SSL}}  & \multicolumn{2}{c|}{\textbf{Pool-5f}}   & \multicolumn{2}{c|}{\textbf{Pool-11f}} & \multicolumn{2}{c|}{\textbf{Pool-21f}} & \multicolumn{2}{c|}{\textbf{Pool-Aug}}  \\
  
%    \cmidrule(l{0.0em}r{4.0em}){2-3}  \cmidrule(r{4.0em}){4-5} \cmidrule(r{0.5em}){6-7}
     & \textbf{Metric} $\to$  &  & \textbf{mIoU }  & \textbf{AUC }  & \textbf{mIoU }  & \textbf{AUC }  & \textbf{mIoU }  & \textbf{AUC } & \textbf{mIoU }  & \textbf{AUC }\\
 [0.5ex] 
 \hline
S & Random    & \xmark             & 47.58          & 18.69         & 44.61             & 17.76             & 44.52             & 17.67             & 43.80             & 17.15 \\ 
S & Entropy   & \xmark             & 49.96          & 19.48         & 47.43             & 18.52             & 46.08             & 18.21             & 44.51             & 17.33 \\
S & EqualAL  & \xmark             & 49.50          & 19.29         & 47.14             & 18.44             & 46.32             & 18.18             & 44.24             & 17.29 \\
B & Coreset & \xmark           & 50.08          & 19.44         & 47.72             & 18.69             & 46.68             & 18.38             & 44.70             & 17.54 \\
S & Random-SSL & \cmark            & 47.92          & 19.03         & 45.25             & 18.02             & 46.27             & 18.19             & 44.17             & 17.29 \\
S & Entropy-SSL  & \cmark          & 48.78          & 19.31         & 47.53             & 18.56             & 46.93             & 18.43             & 44.50             & 17.47 \\
S & EqualAL-SSL & \cmark         & 48.80          & 19.28         & 46.50             & 18.39             & 47.11             & 18.54             & 44.81             & 17.56 \\
B & Coreset-SSL  & \cmark           & \textbf{50.44} & \textbf{19.69}& \textbf{48.99}    & \textbf{19.01}    & \textbf{47.62}    & \textbf{18.69}    & \textbf{45.81}    & \textbf{17.74} \\
- & 100\%       & \xmark           & 53.25          & 21.30         & 48.85             & 19.54             & 49.23             & 19.69             & 46.03             & 18.41 \\          
\hline
\multicolumn{11}{|l|}{\textit{With MC-Dropout decoder} }\\
\hline
S & BALD    & \xmark        & 50.40 & 19.29         & 47.85             & 18.74             & 46.78             & 18.57             & 45.53    & 17.80 \\
S & BALD-SSL    & \cmark           & 50.33          & 19.62         & 47.34             & 18.61             & 47.06             & 18.57             & 45.16             & 17.72 \\
%B & Coreset-MCD             & \textbf{50.40} & 19.49         & 47.67             & 18.61             & 46.86             & 18.35             & 44.74             & 17.50  \\
%B & Coreset-MCD-SSL         & 50.28          & \textbf{19.65}& \textbf{48.60}    & \textbf{18.96}    & \textbf{47.73}    & \textbf{18.75}    & 45.37             & 17.75  \\
- & 100\%-MCD       & \xmark        & 53.82          & 21.53         & 50.86             & 20.34             & 50.43             & 20.17             & 46.62             & 18.65 \\
\hline 
\end{tabular}

\caption{Active Learning results on A2D2-Pool5f, A2D2-Pool11f, A2D2-Pool-21f, and A2D2-PoolAug. AUC@50 and mIoU@30 metrics are reported. S and B denotes the single-sample and batch-based acquisition, respectively. }
\label{table:a2d2_redundant}
\end{table*}

\noindent \textbf{Evaluation scheme.}
We evaluate the methods across different data budget settings, denoted by $\mathcal{I}-\mathcal{S}$, where $\mathcal{I}$ is the initial label budget, $\mathcal{S}$ is the sampling-label budget, and $\mathcal{I}$, $\mathcal{S}$ indicates the percentage of the dataset size. Images are sampled randomly to fulfill the initial label budget. For the subsequent steps, images are sampled using the AL acquisition function up to the allowed sampling-label budget. We test datasets with 10-10, 5-5, and 2-2 budget settings.

\begin{figure*}[t!]
\centering
\begin{tabular}{r@{\hspace{1mm}}c}
 \begin{subfigure}{0.45\textwidth}
  \includegraphics[trim={0 0.5cm 0 0cm}, clip, width=65mm]{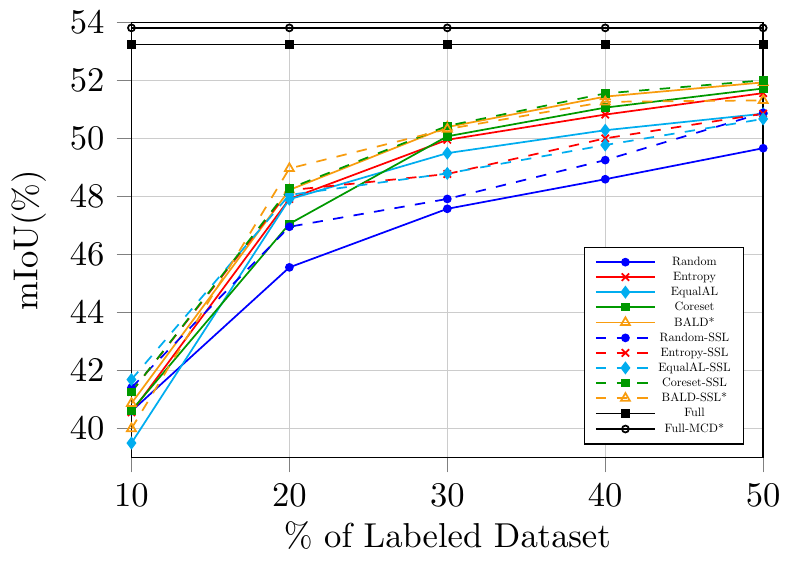}  \caption{A2D2:Pool-5f}
  \end{subfigure} 
  &
 \begin{subfigure}{0.45\textwidth}
 \includegraphics[trim={0 0.5cm 0 0cm}, clip, width=65mm]{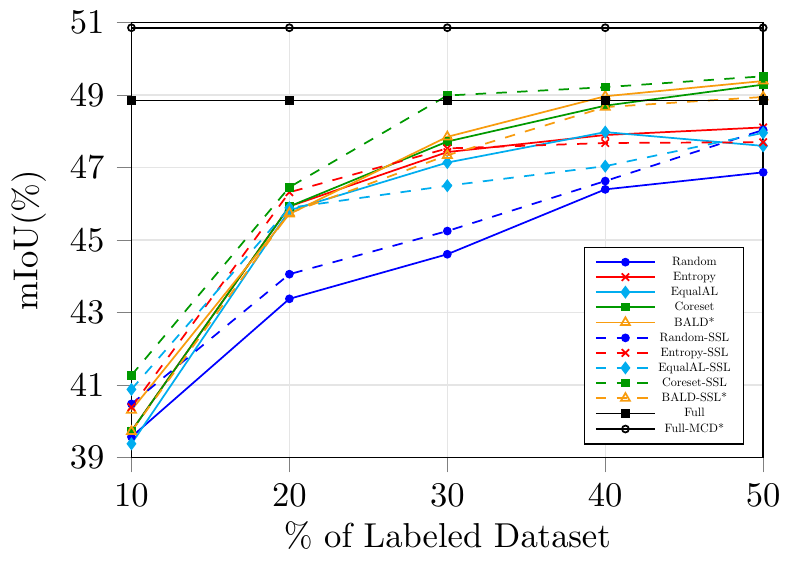} \caption{A2D2:Pool-11f}
  \end{subfigure} \\
   \begin{subfigure}{0.45\textwidth}
  \includegraphics[trim={0 0.5cm 0 0cm}, clip, width=65mm]{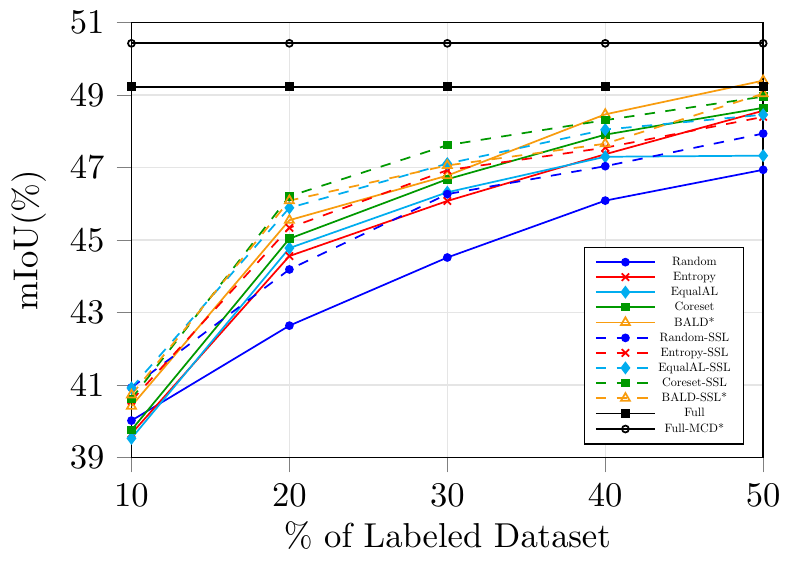} \caption{A2D2:Pool-21f}
   \end{subfigure}  
&     \begin{subfigure}{0.45\textwidth}
 \includegraphics[trim={0 0.5cm 0 0cm}, clip, width=65mm]{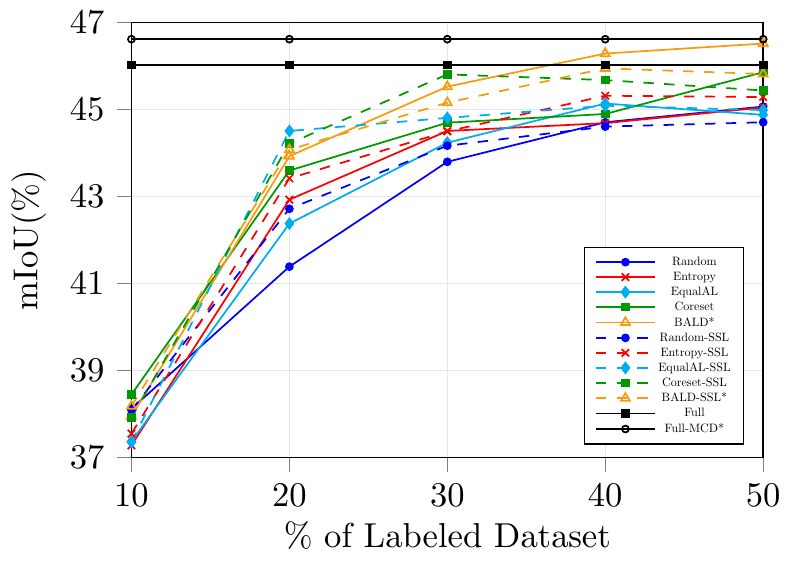} \caption{A2D2:Pool-Aug}
   \end{subfigure} 
\end{tabular}
\caption{Results on redundant datasets. Active Learning performance curves on A2D2 dataset: Pool-5f, Pool-11f, Pool-21f, and Pool-Aug. The X-axis shows the percentage of labeled datasets. The methods which utilize MC-Dropout in their network architecture are marked with $\ast$.}
\label{fig:a2d2_pools_}
\end{figure*}

\noindent \textbf{Evaluation metrics.} We use mean Intersection over Union (mIoU) to evaluate the performance of the model at each AL cycle step. For the evaluation of the active learning method, we use two metrics: Area Under the Budget Curve (AUC@B) and mean Intersection over Union at a budget B (mIoU@B). \textbf{AUC@B} is the area under the performance curves, shown in Figure~\ref{fig:AL_diverse} and \ref{fig:a2d2_pools_}. It captures a cumulative score of the AL performance curve up to a budget B, where B is the percentage of the labeled dataset size. For the experiments on A2D2 pools, we use B=50 in the 10-10 setting. For PASCAL-VOC, we run three experiments with B=10, 25, and 50 in 2-2, 5-5, and 10-10 settings, respectively. For  Cityscapes, we experiment with B=50 in the 10-10 setting. \textbf{mIoU@B} reports the performance of the model after using a certain labeling budget B. We report performance at an intermediate labeling budget to clearly see the ranking of the AL methods.

%----------------------------------------------------------------------

\section{Results}\label{sec:results}

Here, we answer the three questions raised in Section~\ref{sec:intro} concerning the behavior of active learning methods w.r.t data distribution in terms of redundancy, integration of semi-supervised learning, and different labeling budgets. For each experiment, we compare random sampling, single-sample, or batch-based acquisition approaches. 

\subsection{Impact of Dataset Redundancy}~\label{sec:res:data}
%\subsection{Diverse data needs Single-sample method, Redundant needs Batch-based method}~\label{sec:res:data}
%\subsection{Does Data Redundancy Impact AL method Selection?}
%Dataset Redundancy can change the strategy.

Table~\ref{table:sota_results_cityscapes} and Figure~\ref{fig:AL_diverse} show the results on Cityscapes and A2D2 Pool-0f. For both datasets, the single-sample (S) method, EqualAL, performs the best in the supervised-only setting. Table~\ref{table:sota_results_pasca_voc} shows the results obtained on the PASCAL-VOC dataset in 5-5 and 10-10 settings. Single-sample-based methods perform the best in the 10-10 setting, whereas Coreset performs the best in the 5-5 AL setting by a marginal gap w.r.t. random baseline.
Table~\ref{table:a2d2_redundant} and Figure~\ref{fig:a2d2_pools_} show the results for the redundant datasets. The batch-based Coreset method consistently performs the best in all four datasets in the supervised-only setting.
%, whereas there is no consistent winner method in the MCD setting.

\textbf{Diverse datasets need a single-sample method and redundant datasets need a batch-based method.} 
We observe that the order of best-performing models changes based on the level of redundancy in the dataset. Single-sample-based acquisition functions perform best on diverse datasets, whereas batch-based acquisition functions perform best on redundant datasets.
We attribute this reversed effect to the mode collapse problem, where, for redundant datasets, single-sample acquisition methods select local clusters of similar samples. Diverse datasets are devoid of this issue as they do not possess local clusters due to high diversity across samples. Therefore, the diversity-driven acquisition is not critical for diverse datasets. 

This observation is consistent for PASCAL-VOC, where single-sample-based uncertainty-type methods perform better than batch-based and random methods in the high-budget setting. The difference between the methods is only marginal here since most acquired samples add ample new information due to the highly diverse nature of the dataset. This difference further diminishes w.r.t. random baseline with a lower labeling budget (\textit{e.g.} 5-5) since any learned useful bias also becomes weaker. The observations for the 5-5 setting tend towards a very low-budget setting which is further analysed in Section~\ref{sec:res:low}. 
%We argue this is because the dataset is highly diverse; therefore, every acquired sample brings new information.

\paragraph{Mode collapse analysis.} 
Here, we analyse and visualize the above mentioned model collapse issue. Mode collapse in active learning refers to the circumstance that acquisition functions tend to select a set of similar (redundant) samples when acquiring batches of data \cite{Kirsch2019BatchBALD}. 
This can occur when a single-sample acquisition function gives a high score to at least one of the similar samples in the set. Since similar samples have highly overlapping information, all samples in the set receive a high score. Thus, all similar samples tend to be selected, causing this collapse.
Since the selected samples are all very similar, their annotation does not add much more value to the model than if a single sample was added. 

\begin{figure*}[t!] 
\centering
\begin{tabular}{c@{\hspace{1mm}}c@{\hspace{1mm}}c}
\begin{subfigure}{0.25\textwidth}
 \includegraphics[trim={0 0cm 0 0cm}, clip, width=40mm]{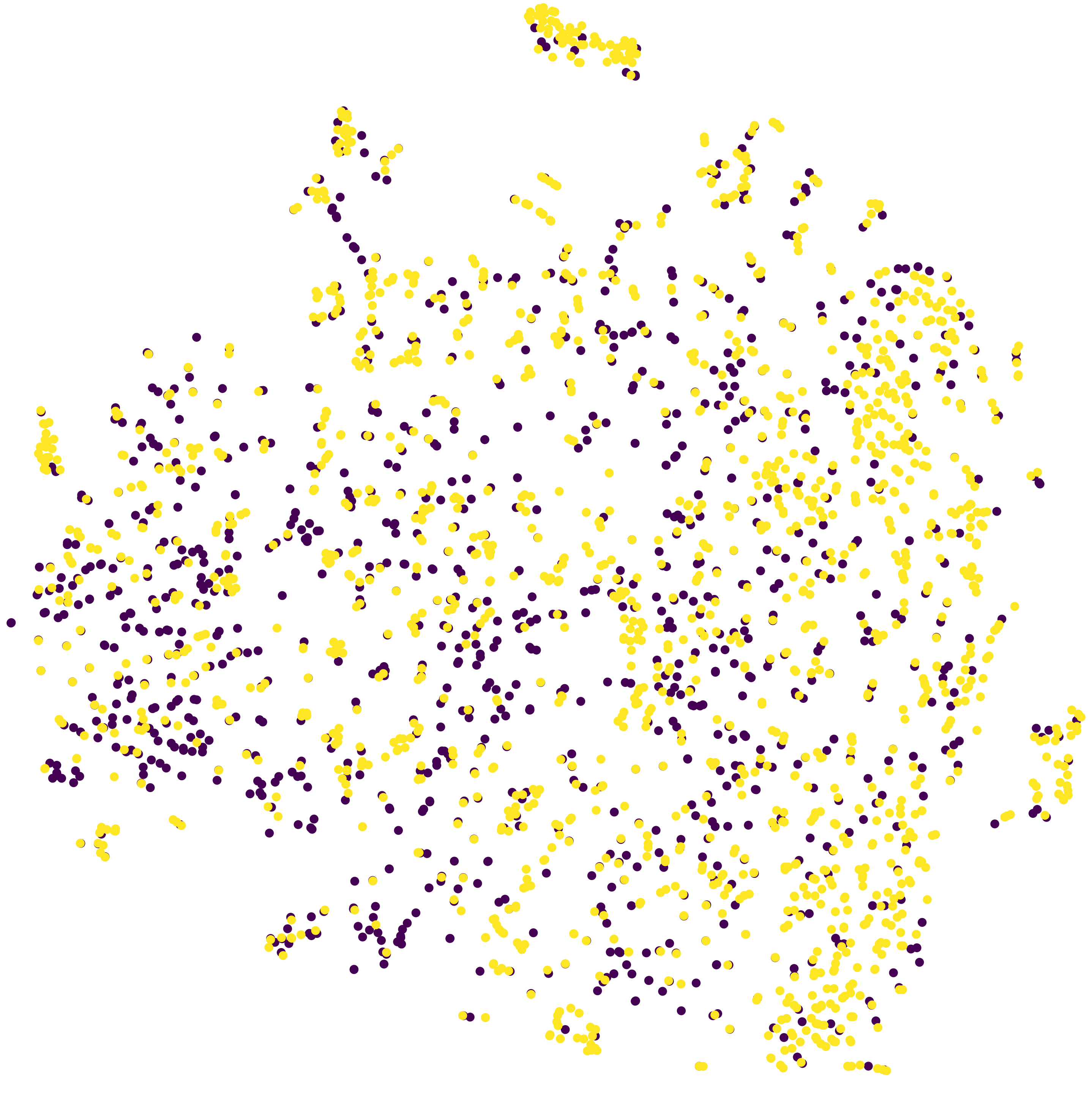} \caption{Coreset (Batch-based AL)} 
 \end{subfigure} &
 \begin{subfigure}{0.25\textwidth}
\includegraphics[trim={0 0cm 0.3cm 0.28cm}, clip,width=40mm]{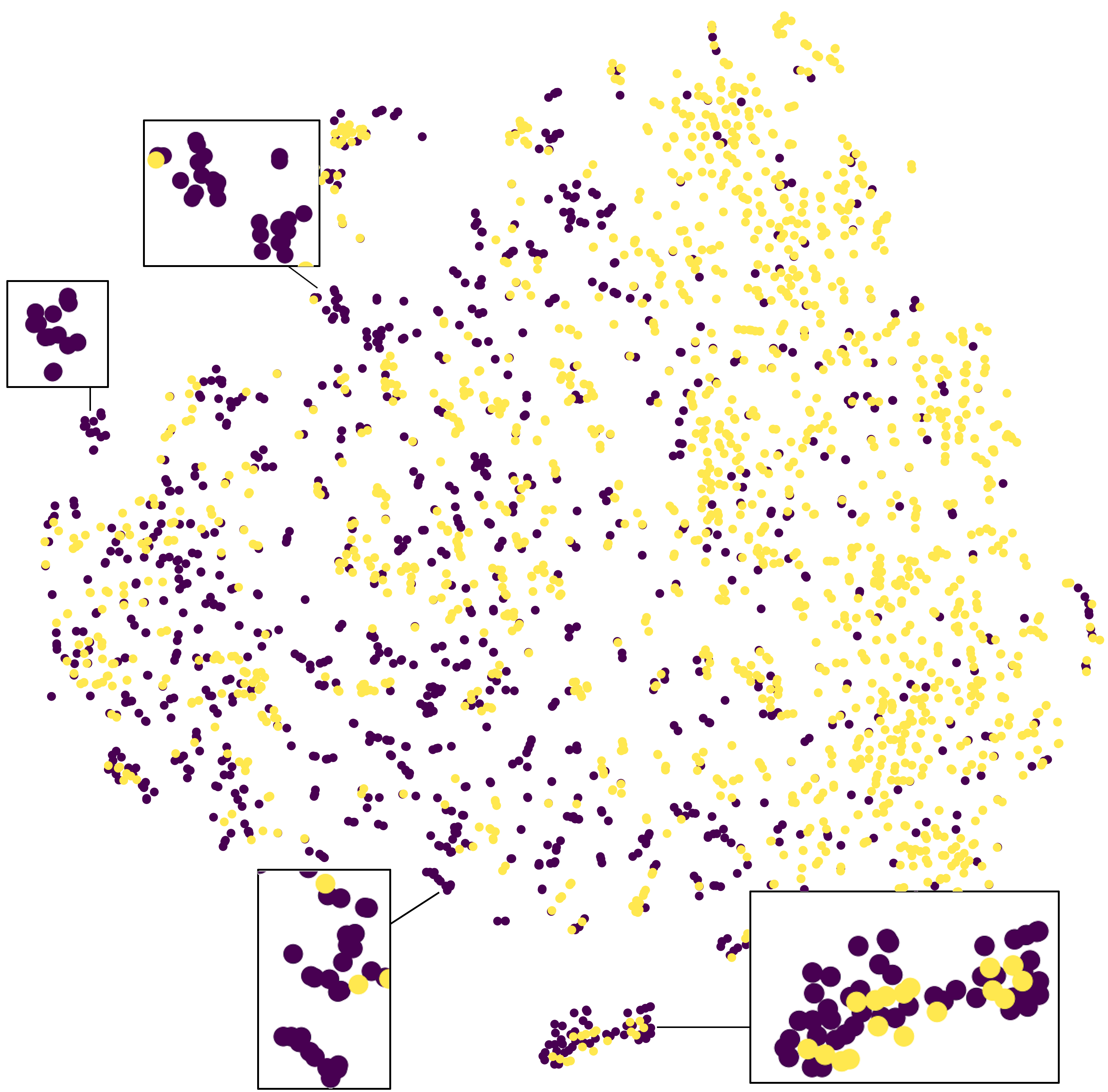}  \caption{Entropy (Single-sample AL)}
  \end{subfigure} &
   \begin{subfigure}{0.4\textwidth}
 \includegraphics[trim={0 0cm 0 0cm}, clip, width=65mm]{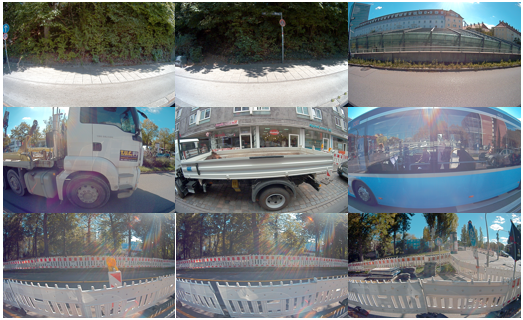} \caption{Mode collapse (Single-sample AL)}
  \end{subfigure}
\end{tabular}
\caption{TSNE plots of (a)~Coreset and (b)~Entropy functions for A2D2 Pool-21f.   
The yellow points are feature representation from the unlabeled set, the violet point are the acquired points. The batch-based approach has good selection coverage, whereas the single-sample acquisition approach selects similar samples from clusters. Figure (c) shows acquired redundant samples from the violet clusters in (b).}
\label{fig:TSNE}
\end{figure*}

We provide a qualitative analysis of the mode collapse issue on the redundant A2D2 Pool-21f. We plot the feature representations using t-SNE to show the selection process for a single-sample-based Entropy function and batch-based Coreset function, shown in Figure~\ref{fig:TSNE}.
It shows that Entropy acquisition selects many samples within local clusters, which are similar samples with overlapping information. This yields a suboptimal use of the annotation budget. In contrast, Coreset acquisition has a good selection coverage and avoids this mode collapse.

In this work, we argue that mode collapse is a common issue in many real-world datasets, containing similar samples. A good acquisition function for such datasets must be aware of the batch's diversity to address the mode collapse issue. It is largely ignored due to the narrow scope of existing AL benchmarks like PASCAL-VOC and Cityscapes, which only cover diverse datasets.

% We provide a qualitative analysis of the mode collapse issue on the redundant dataset A2D2 Pool-21f. We plot the feature representations using t-SNE to show the selection process for a single-sample-based entropy acquisition function and batch-based Coreset function. It shows that entropy acquisition selects many samples within local clusters, which are similar samples with overlapping information. Hence, we can observe the mode collapse effect. In contrast, Coreset acquisition has a good selection coverage while avoiding mode collapse.

\subsection{Systematic Integration of SSL}\label{sec:res:SSL}

%\tb{The following is hard to digest. Can this be structured a bit more wrt conclusion?}

%The Coreset-SSL performs best in all redundant datasets, see Table~\ref{table:a2d2_redundant} and Figure~\ref{fig:a2d2_pools_}.

For all redundant datasets, the Coreset-SSL approach consistently performs the best; see results in Table~\ref{table:a2d2_redundant} and Figure~\ref{fig:a2d2_pools_}.
For diverse datasets, SSL integration is also helpful, but there is no consistent best approach. For the PASCAL-VOC dataset, single-sample based methods with SSL show the best performance, shown in Table~\ref{table:sota_results_pasca_voc}. For Cityscapes, Coreset-SSL outperforms all other approaches; see Table~\ref{table:sota_results_cityscapes} and Figure~\ref{fig:AL_diverse}. For A2D2-Pool0f, Coreset-SSL improves over Coreset, but the single-sample acquisition method BALD approach shows the best performance.

%Table~\ref{table:sota_results_pasca_voc} shows the results when integrating semi-supervised learning on the PASCAL-VOC dataset in 5-5 and 10-10 settings. EqualAL-SSL and Entropy-SSL, which are single-sample based methods, show the best performance in both settings.
%For Cityscapes, Coreset with SSL performs best, outperforming all the standard methods, see Table~\ref{table:sota_results_cityscapes} and Figure~\ref{fig:AL_diverse}. For the A2D2 Pool-0f setting, the Coreset-SSL method improves over Coreset. However, EqualAL and BALD are the best methods in corresponding MCD settings. 
 
%We notice a similar trend for the MCD setting: Coreset-MCD-SSL outperforms BALD with one minor exception for the Pool-Aug.

\textbf{Redundant datasets favour the integration of batch-based active learning and semi-supervised learning.} 
The batch-based acquisition function Coreset always profits from the integration of SSL. 
%In contrast, integration of SSL with single-sample acquisition is either ineffective for most of the datasets, except for PASCAL-VOC. 
Coreset aligns well with the SSL objective since Coreset selects samples from each local cluster, thus covering the whole data distribution. This assists SSL in obtaining maximum information from the unlabeled samples, as discussed in Section~\ref{sec:ALforSemSeg:SemiSuper}.
This effect is especially strong in the redundant A2D2 pools, where Coreset-SSL always improves over Coreset and also shows the best performance. In contrast, SSL integration for single-sample methods is either harmful or ineffective, except for the PASCAL-VOC dataset. 
Interestingly, in Pool-11f, some Coreset-SSL methods even outperform the 100\% baseline with less than 30\% labeled data. This indicates that some labeled redundant samples can even harm the model (see Figure~\ref{fig:a2d2_pools_}), possibly due to data imbalance. 
%\tb{Or do you have a better explanation?} \sud{What do you mean by data imbalance? It could also occur due to overfitting on similar samples, affecting generalization.}
For Cityscapes, SSL with Coreset yields significant improvement, and SSL even changes the ranking of the methods. We see that EqualAL performs the best in the supervised-only setting, whereas Coreset-SSL surpasses all methods. This slight anomaly in the case of Cityscapes happens because the advantage due to the combination of SSL and batch-based method is greater than the advantage of using single-sample methods in non-redundant datasets.
For diverse PASCAL-VOC, all methods align well with SSL. All methods perform well with no clear winner method since all selection criteria select samples that provide good coverage of the data distribution.

%For PASCAL-VOC, all methods align well with SSL integration due to its highly diverse nature.
% of the dataset. 

\subsection{Low Annotation Budget} \label{sec:res:low}
\noindent \textbf{Active learning is volatile with a low budget.}
Experimenting with PASCAL-VOC in the 2-2 budget setting, Random-SSL performs the best, i.e., semi-supervised learning without active learning component. We believe that active learning fails in this setting because it fails to capture any helpful bias for selection in such a low-data regime with diverse samples. Our observations in this low-budget setting confirm and provide a stronger empirical support for similar behavior observed in~\cite{Mittal2019Illusions}. 
For A2D2 Pool-0f and Cityscapes in the 2-2 setting, the single-sample acquisition performs the best, while its SSL integration is detrimental. These methods possibly learn some useful bias due to the specialized driving domain.  
For redundant datasets in low budget settings, batch-based acquisition is still the most effective way. However, SSL does not contribute any additional improvements due to insufficient labeled samples to support learning from unlabeled samples. 
Overall, we observe a highly volatile nature of active learning in conjunction with a low budget. The ideal policy transitions from random selection towards the batch-based acquisition, as the dataset redundancy goes from low to high. 
%In additional experiments on the A2D2 Pool-0f in the 2-2 setting, we find that active learning can be useful again due to a more focused, narrower driving dataset domain. 
The result tables corresponding to this section are included in the Appendix.

% -----------------------------------------------------------------

\begin{table}[!t]
\centering
\begin{tabular}{|c|l@{\hskip4pt}|@{\hskip4pt}c@{\hskip4pt}|@{\hskip4pt}c@{\hskip4pt}|@{\hskip4pt}c@{\hskip4pt}|}
        \hline
      \textbf{A} & \textbf{AL Method} & \textbf{SSL}   & \textbf{mIoU}  & \textbf{AUC} \\
      \hline
      B & Uniform   & \xmark     & 57.75             & ---    \\
      S & Random    & \xmark    & 56.14             & 5.35  \\
      S & Entropy   & \xmark    & 60.16             & 5.53  \\
      B & Coreset   & \xmark    & 60.30             & 5.55  \\
      S & Uniform (@5) + Entropy  & \xmark  & 60.40         & 5.66  \\
      \hline
      B & Uniform-SSL  & \cmark  & 58.93             & ---  \\
      S & Random-SSL  & \cmark    & 57.57             & 5.53  \\
      S & Entropy-SSL  & \cmark  & 59.91             & 5.61  \\
      B & Coreset-SSL   & \cmark  & \textbf{61.13}   & \textbf{5.72} \\
      S & Uniform (@5) + Ent-SSL  & \cmark  & 59.63   & 5.59 \\
      \hline
      - & 100\%     &  \xmark   & 66.65             & 6.64  \\
        \hline
\end{tabular}
% \vspace{30px}
\caption{AL results on the proposed A2D2-3k task. mIoU@7.5 and AUC@7.5 is reported. S and B denotes the single-sample and batch-based acquisition, respectively. Uniform refers to temporal subsampling selection process and (@5) means every $5^{th}$ frame.}
\label{tab:a2d2_3k}
\end{table}

\subsection{An exemplar case study: A2D2-3K task}\label{sec:res:a2d2_3k}

Previous active learning works on semantic segmentation cover only the combination of a diverse dataset and a high annotation budget. In contrast, the collected raw data can be quite redundant, like in video datasets. 
To study this missing redundant setting, we propose a new active learning task A2D2-3K for segmentation based on the A2D2 dataset.
The aim of the new task is to select 3K images (similar size to Cityscapes) from the original A2D2 dataset~($\sim$40K images) to achieve the best performance. We select 3K images using active learning in 3 cycles with 1K images each.
We compare 5 acquisition functions, including Random, Entropy, and Coreset, along with SSL integration.
Such video datasets are often manually subsampled based on some prior information like time or location, and then used for active learning. Therefore, we also include two such baselines - (a) where 3K samples are uniformly selected based on time information, denoted as Uniform, and (b) where every fifth sample is first selected uniformly to select $\sim8K$ samples and then applied with Entropy acquisition function, denoted as Uniform(@5)+Entropy. The second approach is closer to previously used active learning benchmarks in the driving context. 
Results are shown in Table~\ref{tab:a2d2_3k}. We find that the batch-based Coreset-SSL method performs the best, discussed in Section~\ref{sec:res:SSL}, while the subsampling-based approaches are sub-optimal. This makes an excellent case for active learning in datasets with high redundancy, as active learning filters the data better than time-based subsampling methods.  %\sud{add table 6 reference.} \sud{check all captions.} 

%We find that such intuitive approaches are also sub-optimal, whereas the batch-based Coreset-SSL method performs the best as also discussed in Section~\ref{sec:res:SSL}.

%To study one of the missing realistic setting where the raw dataset can be quite redundant like in video datasets, we propose a new task for semantic segmentation based on the A2D2 dataset. 
%A2D2 contains sequence data close to raw data that one would record during a measurement campaign. 

%The aim of the AL task is to select 3K images (similar size to Cityscapes) from the original A2D2 dataset to achieve the best performance. We select 3K images using AL methods in 3 cycles with 1K images each. 
%We compare 5 acquisition functions, including the standard subsampling process. We also evaluate each method along with their SSL setting. 
%We also test another intuitive approach, where the dataset is sub-sampled using prior information like time stamps and then used for active learning based selection. 

%We find that such intuitive approaches are also sub-optimal, whereas the batch-based Coreset-SSL method performs the best as also discussed in Section~\ref{sec:res:SSL}. This confirms with our conclusion and provides a strong case for our proposed guide.
%which is often used to create dataset 

%We find that the batch-based Coreset-SSL method performs the best, similar to other redundant dataset experiments (see Table~\ref{tab:a2d2_3k}). 
% Results are shown in Table~\ref{tab:a2d2_3k}.

% \input{5_conclusions}
\section{Conclusion}
\begin{table}[!t]
\centering
\setlength{\tabcolsep}{3pt}
\begin{tabular}{|@{\hspace{1mm}}l@{\hspace{1mm}}|c|c|c|c|}
        \hline
       \multirow{2}{*}{Dataset $\downarrow$} & \multicolumn{4}{c|}{Annotation Budget}\\
        & \multicolumn{2}{c|}{Low} & \multicolumn{2}{c|}{High} \\ 
        \cline{2-5}
      Sup. $\rightarrow$ & AL  & SSL-AL & AL  & SSL-AL\\
      \hline
      Diverse      & \cellcolor{green!25}Random       & \cellcolor{green!25}Random-SSL  & \cellcolor{black!20}Single       & \cellcolor{black!20}Single-SSL   \\
      Redundant    & \cellcolor{green!25}Batch     & \cellcolor{green!25}Batch  & \cellcolor{green!25}Batch    & \cellcolor{green!25}Batch-SSL   \\
      \hline
\end{tabular}
\caption{Overview showing the best performing AL method for each scenario. Single and Batch refer to single-sample and batch-based method, and Random refers to random selection. Suffix -SSL refers to the usage of semi-supervised learning.}
\label{tab:overview}
\end{table}
% This work shows that active learning is indeed a useful tool for semantic segmentation. 
% However, it is vital to understand the essentials for successfully using AL methods in different settings across data distributions, acquisition functions, annotation budgets, and the integration of semi-supervised learning. 
% We find that single-sample-based uncertainty is a suitable measure for selection in diverse datasets, whereas batch-based diversity-driven measures are more suited for redundant datasets. We find SSL always successfully integrates with batch-based diversity-driven methods. However, it can be destructive when combined with single-sample-based uncertainty acquisition functions. 
% Given these findings are missing in the current method development, new methods created w.r.t. the current standards yield methods that are optimized only for special scenarios.
% We find that current segmentation AL benchmarks in the driving scenario are unrealistic and propose a realistic task.
 
This work shows that active learning is indeed a useful tool for semantic segmentation. However, it is vital to understand the behavior of different active learning methods in various application scenarios. 
Table~\ref{tab:overview} provides an overview of the best performing methods for each scenario. Our findings indicate that single-sample-based uncertainty is a suitable measure for sample selection in diverse datasets. In contrast, batch-based diversity-driven measures are better suited for datasets with high levels of redundancy. SSL is successfully integrated with batch-based diversity-driven methods. However, it can have a detrimental impact when combined with single-sample-based uncertainty acquisition functions. Active learning with low annotation budgets is highly sensitive to the level of redundancy in the dataset.
%not effective with low annotation budgets for diverse datasets, and .
These findings have been missing in method development, which are optimized only for a few scenarios. The results of this study facilitate a broader view on the task with presumably positive effects in many applications.  

\section*{Acknowledgement}
The authors would like to thank Philipp Schr\"{o}ppel, Jan Bechtold, and Mar\'ia A. Bravo for their constructive criticism on the manuscript.
The research leading to these results is funded by the German Federal
Ministry for Economic Affairs and Climate Action within the project ``KI
Delta Learning'' (Forderkennzeichen 19A19013N) and ``KI Wissen – Entwicklung 
von Methoden für die Einbindung von Wissen in maschinelles Lernen''. The authors would like to thank the consortium for the successful cooperation. Funded by the Deutsche Forschungsgemeinschaft (DFG) - 417962828.

% \textbf{Checklist}
% \begin{itemize}
%     \item Check paragraph endings for a smooth and meaningful transition to the next paragraph. 
%     \item Check relevancy of each paragraph towards the final goal of the study. Including datasets explanations.
% \end{itemize}

\newpage

% {\small
% \bibliographystyle{ieee_fullname}
% \bibliography{egbib}
% }
% {\small
% \bibliographystyle{ieee_fullname}
% \bibliography{ref}
% }

\clearpage
\newpage

\appendix

\section{Datasets Visualization}
\begin{figure}[t]
\centering
\begin{tabular}{c@{\hspace{1mm}}c@{\hspace{1mm}}c@{\hspace{1mm}}c}
%(a) Original & (b) Ground Truth & (c) Baseline  & (d) Our Results  \\[6pt]

% \vspace{2mm}
\begin{turn}{90}\footnotesize{Cityscapes}\end{turn} &
\includegraphics[width=25mm]{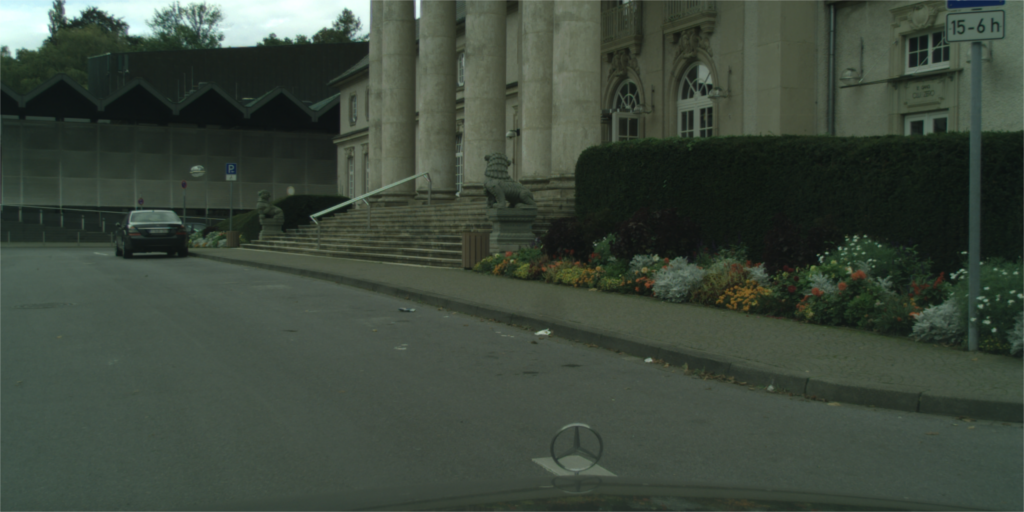} &
\includegraphics[width=25mm]{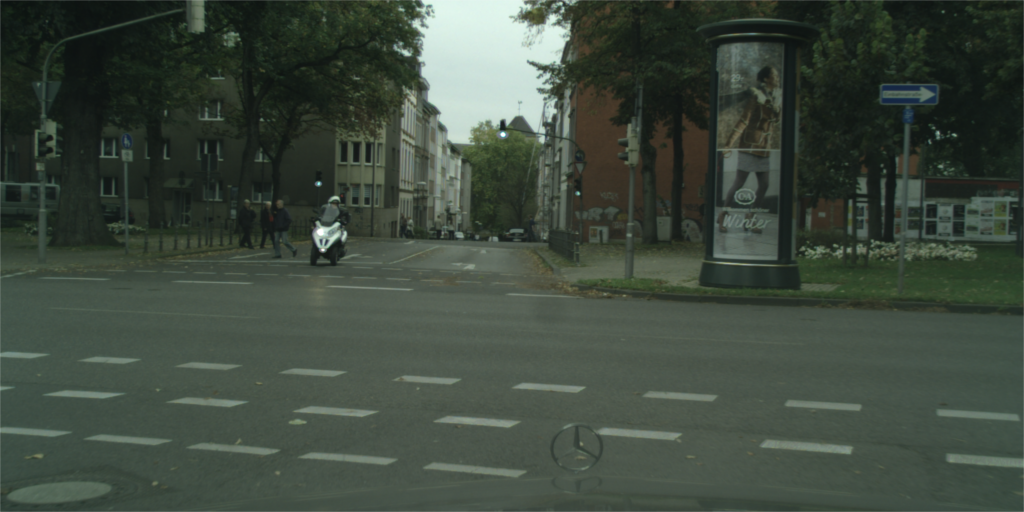} &
\includegraphics[width=25mm]{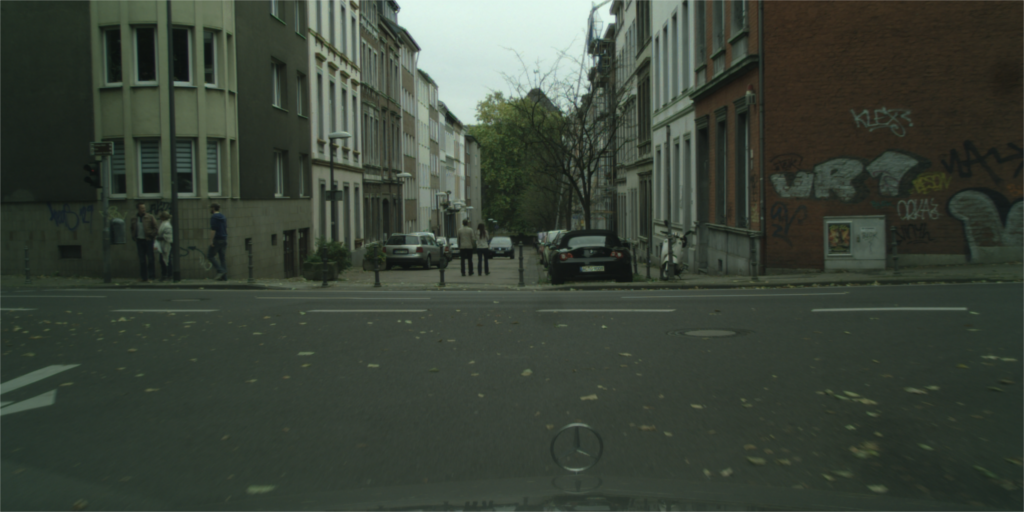} \\
\begin{turn}{90}\footnotesize{Cityscapes}\end{turn} &
\includegraphics[width=25mm]{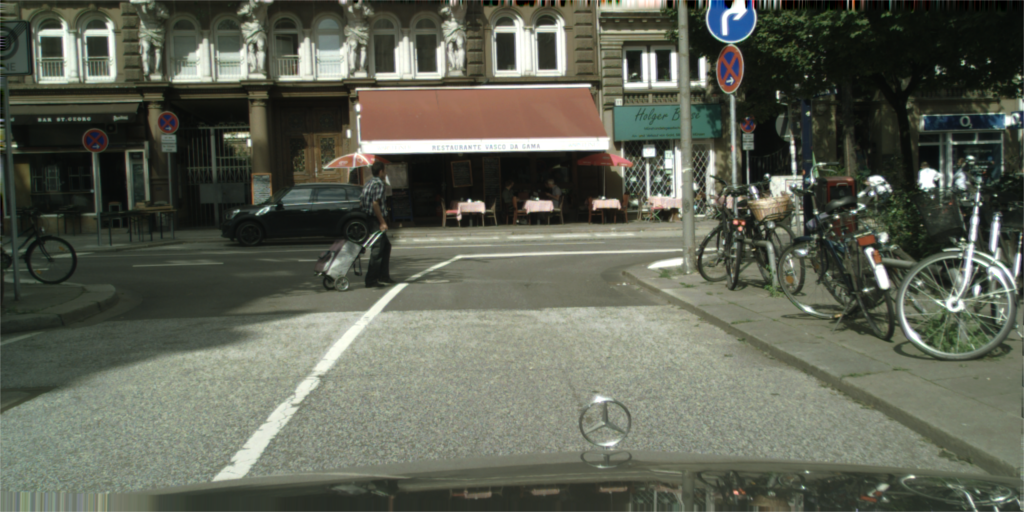} &
\includegraphics[width=25mm]{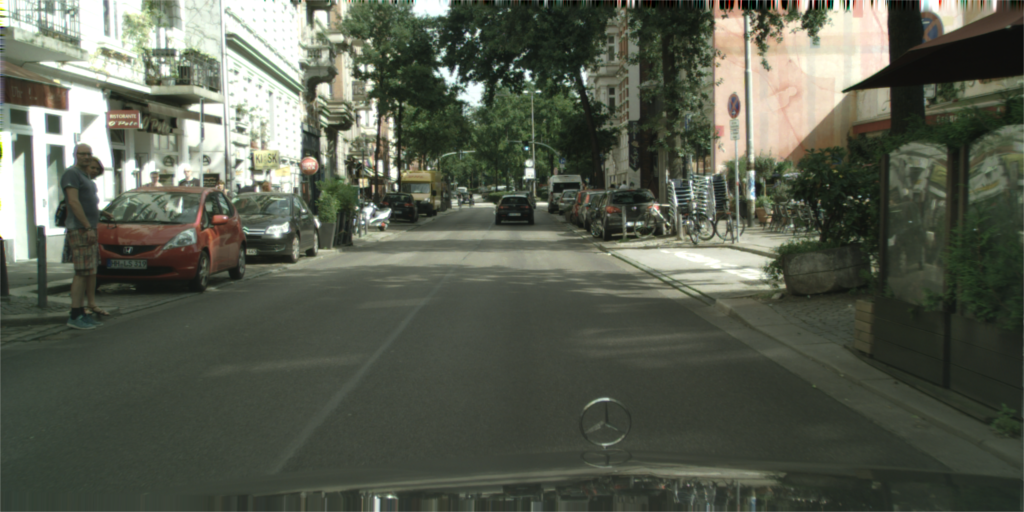} &
\includegraphics[width=25mm]{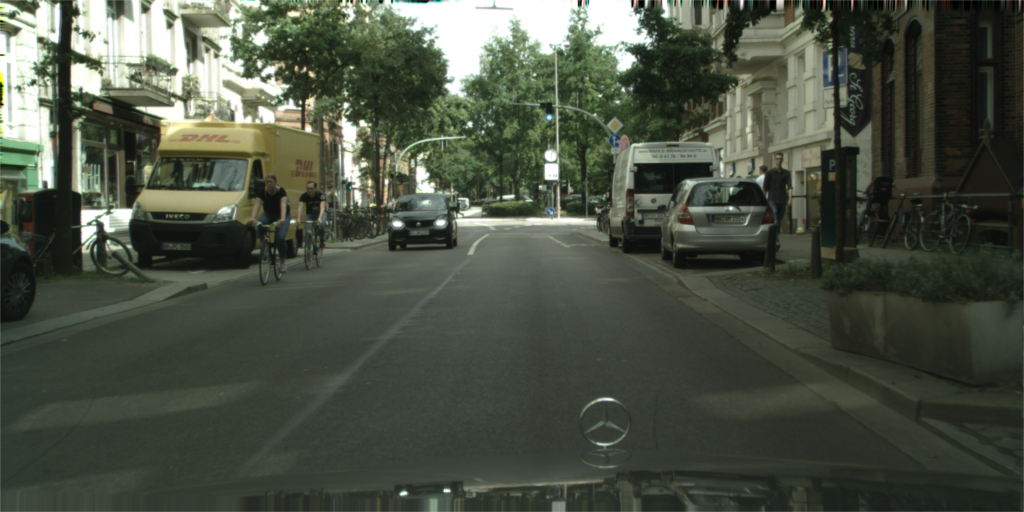} \\
\begin{turn}{90}\hspace{4mm}A2D2\end{turn} &
\includegraphics[width=25mm]{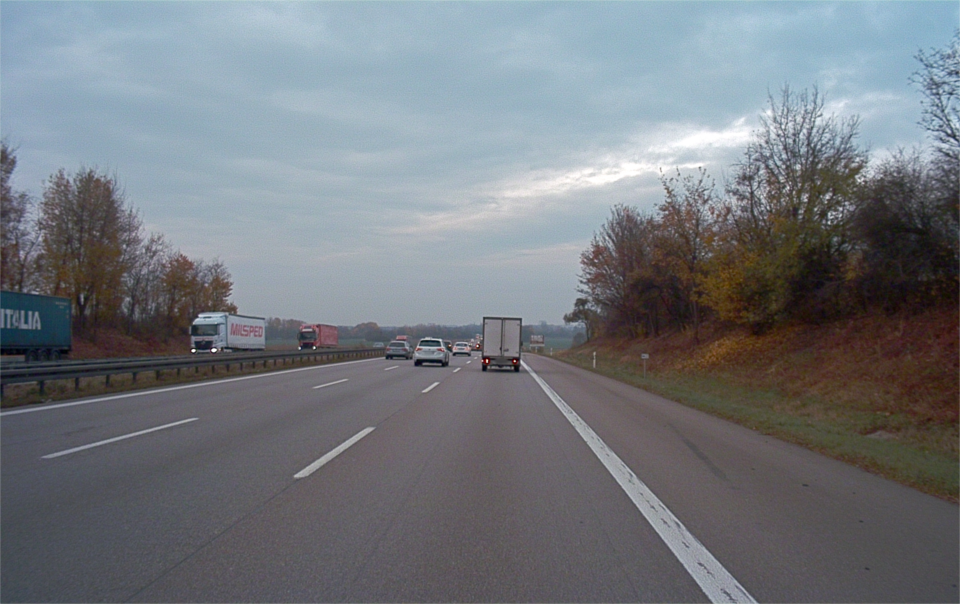} &
\includegraphics[width=25mm]{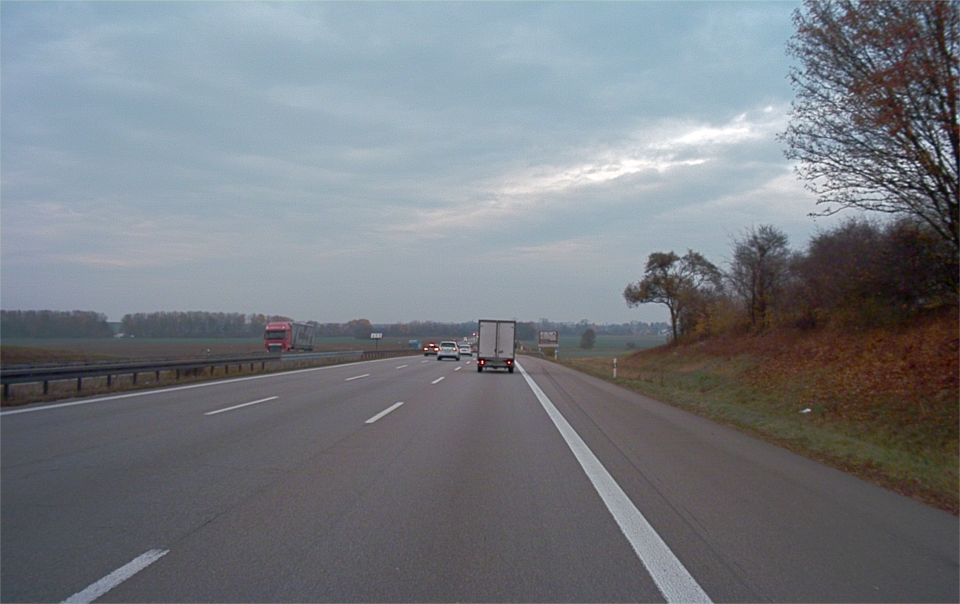} &
\includegraphics[width=25mm]{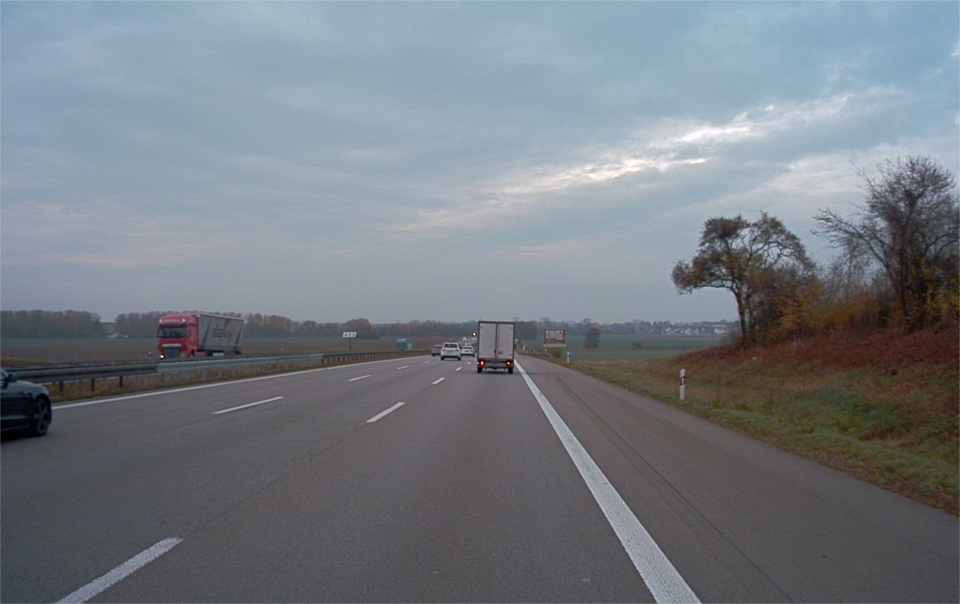} \\
\begin{turn}{90}\hspace{4mm}A2D2\end{turn} &
\includegraphics[width=25mm]{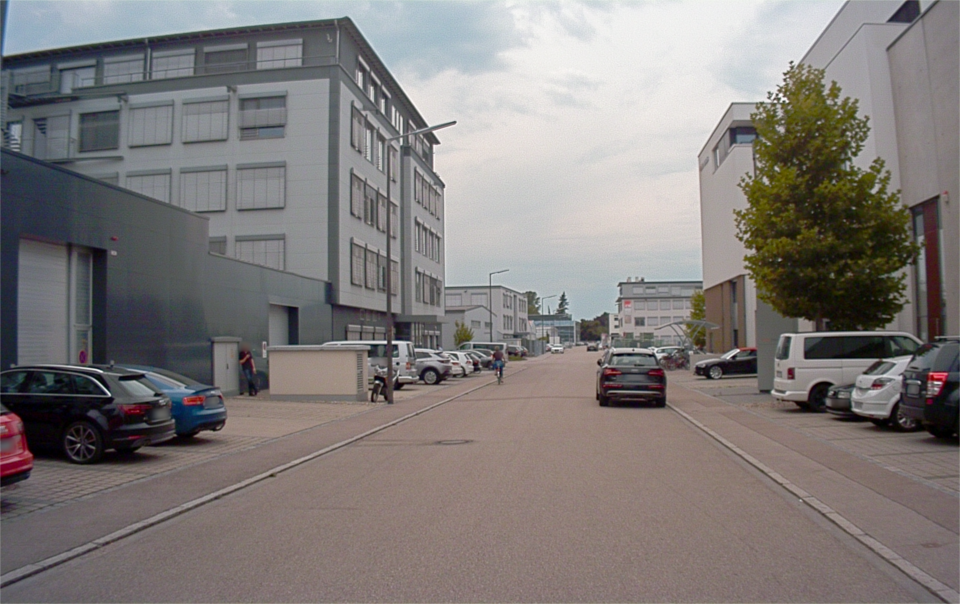} &
\includegraphics[width=25mm]{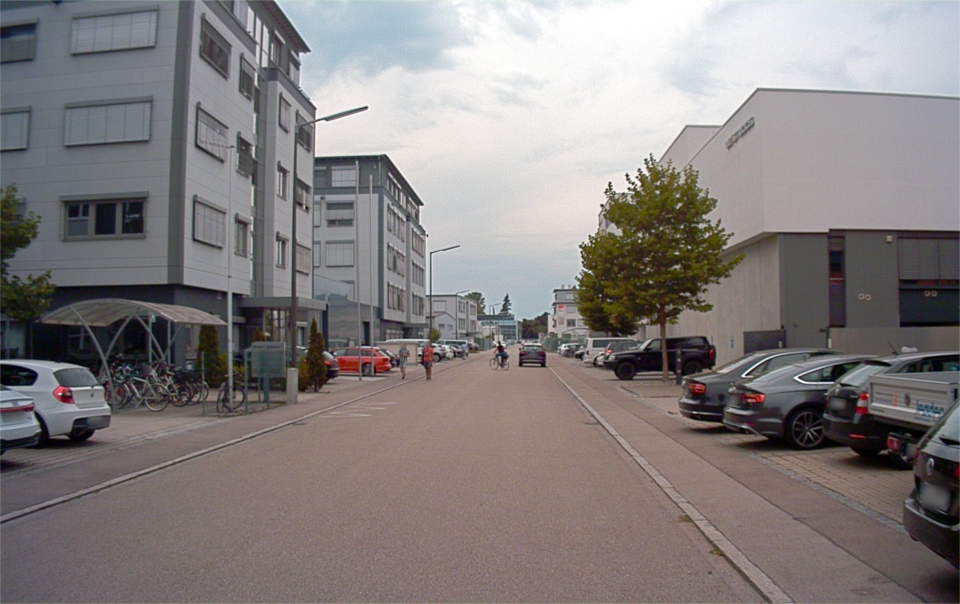} &
\includegraphics[width=25mm]{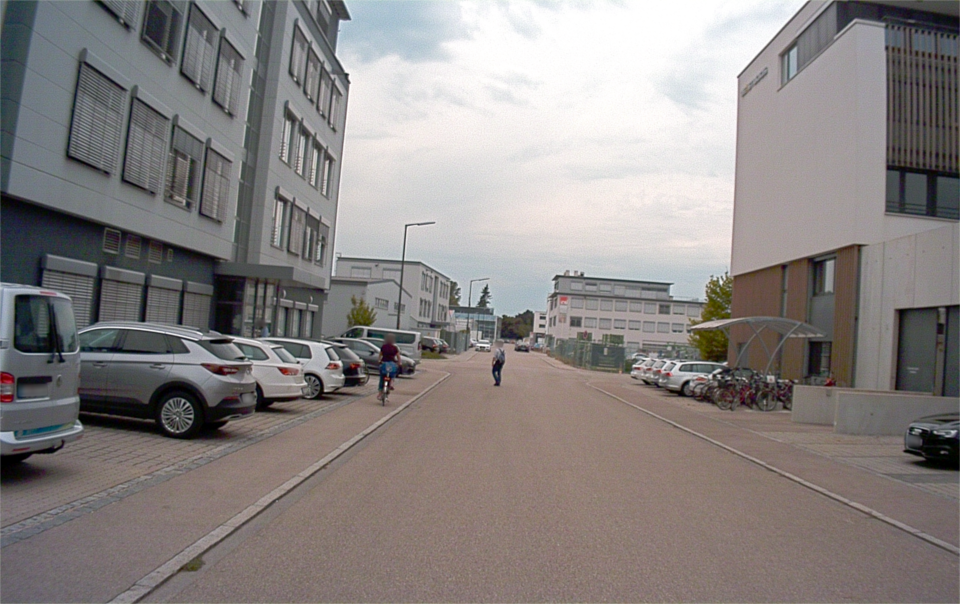} \\
\end{tabular}
\caption{Consecutive images from the Cityscapes and A2D2 datasets. This shows even the consecutive images in the Cityscapes dataset are different and diverse, whereas consecutive frames in the A2D2 dataset are very similar, containing redundant information.}
\label{fig:DatasetExmpl}

\end{figure}

Figure~\ref{fig:DatasetExmpl} shows examples of the A2D2 and the Cityscapes dataset. 
Each row shows three temporally consecutive frames in both labeled datasets. We clearly observe that the images in the A2D2 dataset have high-overlapping information, whereas images in the Cityscapes dataset are quite diverse.
Therefore, to create our redundancy experiments, we chose the A2D2 dataset as the base dataset. %Figure~\ref{fig:Crop_and_augment} shows how A2D2-Pool-Aug is created using augmentation.

%%%%%%%%% BODY TEXT
\section{Training details}
We used the DeepLabv3+~\cite{deeplabv3+} architecture with Wide-ResNet38 (WRN-38) \cite{wide38} backbone for all our experiments. The backbone WRN-38 is pre-trained using ImageNet~\cite{5206848}. For the supervised learning setting, the model is trained using the SGD optimizer with a base-learning rate of $1e-3$, momentum of $0.9$, and a weight decay of $5e-4$. 
We utilize a polynomial learning rate scheduler with a batch size of 8 and train a model in each AL cycle for 100 epochs. The model is trained with data augmentations, including random cropping and random horizontal flipping. Input image size is $256 \times 512$ for Cityscapes and A2D2 datasets and $321 \times 321$ for the PASCAL-VOC dataset.
We utilize the s4GAN~\cite{mittal2019semi} method for semi-supervised learning (SSL). We use the same training setting for the segmentation model as in the supervised setting. 
We use the same hyperparameters as mentioned in \cite{mittal2019semi}, except for the learning rate of the discriminator which is set to $2.5e-5$ for Cityscapes and A2D2 experiments. 
We add 3 dropout layers with a dropout rate of 0.1 in the decoder of the segmentation model for all the MCD-based AL methods.

\section{Evaluation Metric: AUC@B}
We use the following formula to compute the Area Under the Budget Curve(AUC@B) at a total budget B, where B is the percentage of the labeled dataset:

\begin{equation}
   AUC@B = \sum_{i=1}^{i=N} \frac{(b_{i+1}-b_i)(p_i+p_{i+1})}{2}
\end{equation}
,where N is the number of AL acquisition steps, $b_i$ is the percentage of labeled dataset at step $i$, and $p_i$ is the performance of the model in mIoU(\%) at step $i$.

%\clearpage

\begin{figure*}[]
\centering
\begin{tabular}{c@{\hspace{1mm}}c}
\begin{subfigure}{0.45\textwidth}
 \includegraphics[trim={0 0.5cm 0 0cm}, clip, width=65mm]{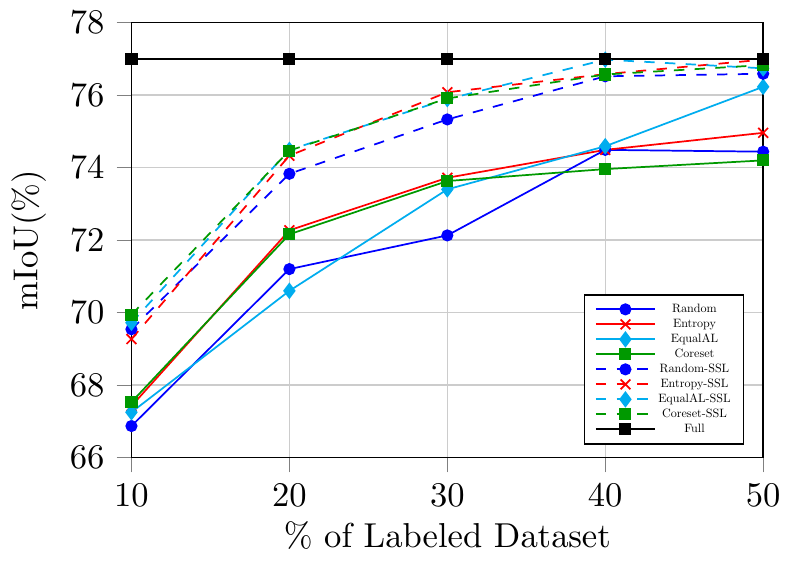}  \caption{PASCAL-VOC: 10-10} \label{fig:pascal_10_10}
 \end{subfigure}  &
 \begin{subfigure}{0.45\textwidth}
  \includegraphics[trim={0 0.5cm 0 0cm}, clip, width=65mm]{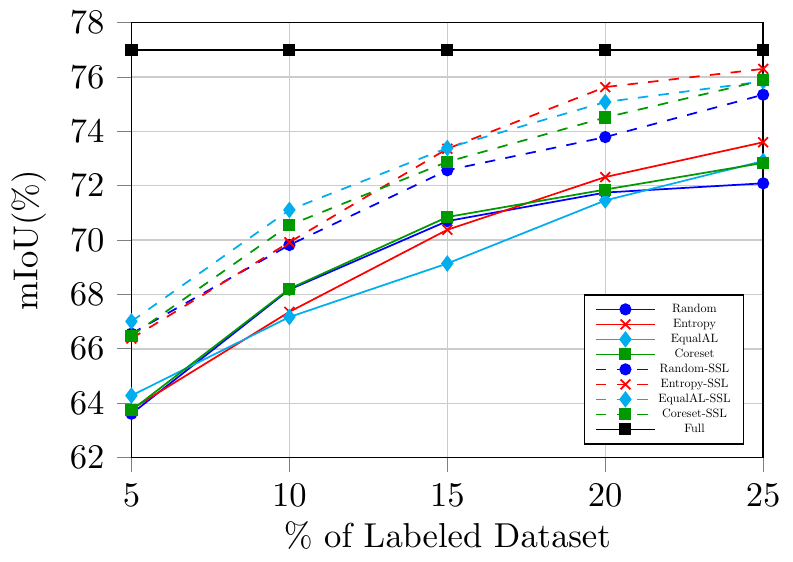} \caption{PASCAL-VOC: 5-5} \label{fig:pascal_5_5}
 \end{subfigure}  \\
  \begin{subfigure}{0.45\textwidth}
  \includegraphics[trim={0 0.5cm 0 0cm}, clip, width=65mm]{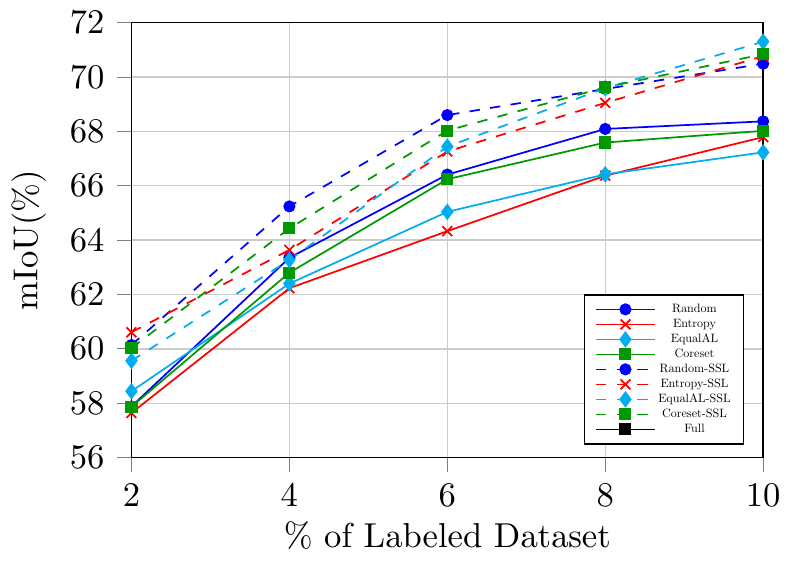} \caption{PASCAL-VOC: 2-2} \label{fig:pascal_2_2} 
 \end{subfigure}  &
 \begin{subfigure}{0.45\textwidth}
  \includegraphics[trim={0 0.5cm 0 0cm}, clip, width=65mm]{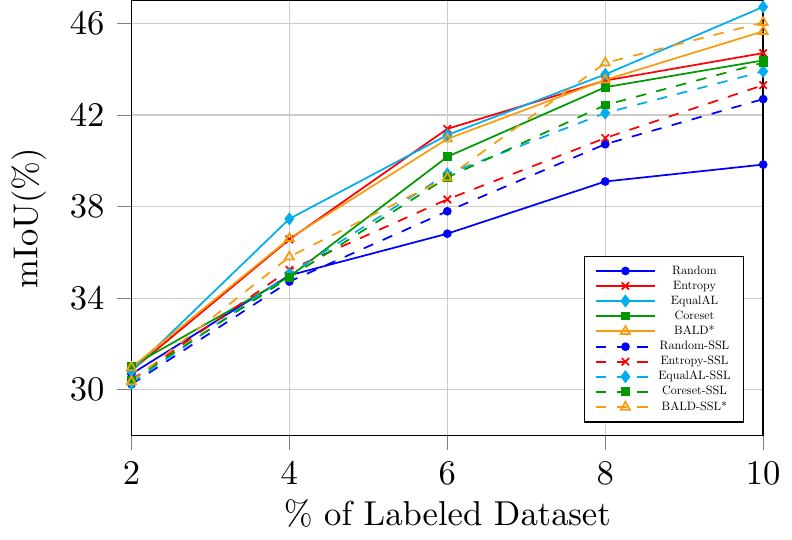} \caption{A2D2: Pool-0f 2-2} \label{fig:a2d2_2_2} 
 \end{subfigure} 
\end{tabular}
\caption{Active learning performance curves on PASCAL-VOC and A2D2:Pool-0f. X-axis shows the percentage of labeled dataset. The methods which utilize MC-Dropout in their network architecture are marked with $\ast$, and are only comparable to other methods with MC-Dropout. }
\label{fig:AL_all}
\end{figure*}

\begin{table}
\centering
\setlength\tabcolsep{3.5pt}
 \begin{tabular}{|c|@{\hspace{1.5mm}}l@{\hspace{1.5mm}}|@{\hspace{0.15cm}}c@{\hspace{0.15cm}}|@{\hspace{0.15cm}}c@{\hspace{0.15cm}}c@{\hspace{0.15cm}}|} 
 \hline 
 \multirow{2}{*}{\textbf{A}} & \textbf{AL Method}   & \multirow{2}{*}{\textbf{SSL}} & \multicolumn{2}{c@{\hspace{1.5mm}}|}{\textbf{PASCAL: 2-2}}   \\  
  
%    \cmidrule(l{0.0em}r{4.0em}){2-3}  \cmidrule(r{4.0em}){4-5} \cmidrule(r{0.5em}){6-7}
    & \textbf{Metric} $\to$ &   & \textbf{mIoU@6}  & \textbf{AUC@10} \\
 [0.5ex] 
 \hline
S & Random        &     \xmark    & 66.41              & 5.22    \\ 
S & Entropy        &    \xmark    & 66.33              & 5.11    \\
S & EqualAL         &  \xmark     & 65.04              & 5.13    \\
B & Coreset         &   \xmark    & 66.24              & 5.19    \\
S & Random-SSL      &   \cmark    & \textbf{68.60}     & \textbf{5.37}    \\
S & Entropy-SSL     &   \cmark    & 67.26              & 5.31    \\
S & EqualAL-SSL     &  \cmark     & 67.44              & 5.31    \\
B & Coreset-SSL     &  \cmark    & 68.03              & 5.35    \\
\hline
- & 100\%           &  \xmark     & 77.00              & 6.16    \\ 
\hline
\end{tabular}

\caption{Active Learning results on PASCAL-VOC dataset in low-budget 2-2 setting. AUC@10 and mIoU@6 metric are reported. A denotes Acquisition method type. S and B denotes the single-sample and batch-based acquisition, respectively.}
\label{table:pascal_2_2}
\end{table}

\begin{table*}[t]
\centering
\setlength\tabcolsep{3.0pt}
 \begin{tabular}{|@{\hspace{0.15cm}}c@{\hspace{0.15cm}}|@{\hspace{0.15cm}}l@{\hspace{0.15cm}}|@{\hspace{0.15cm}}c@{\hspace{0.15cm}}|@{\hspace{0.15cm}}c@{\hspace{0.15cm}}c@{\hspace{0.15cm}}|@{\hspace{0.15cm}}c@{\hspace{0.15cm}}c@{\hspace{0.15cm}}|} 
 \hline 
  \multirow{2}{*}{\textbf{A}} & \textbf{AL Method}     & \multirow{2}{*}{\textbf{SSL}} & \multicolumn{2}{c|@{\hspace{0.15cm}}}{\textbf{A2D2 Pool-0f 2-2}} & \multicolumn{2}{c|}{\textbf{A2D2 Pool-11f 2-2}}    \\
   &   \textbf{Metric} $\to$  & & \textbf{mIoU@6\small}  & \textbf{AUC@10\small} & \textbf{mIoU@6\small}  & \textbf{AUC@10\small}  \\
 [0.5ex] 
 \hline
S & Random      &  \xmark         & 36.82            & 2.92   & 37.74  & 2.93 \\ 
S & Entropy     &   \xmark        & \textbf{41.40}   & 3.18   & 36.37  & 2.92 \\
S & EqualAL     &   \xmark        & 41.13            & \textbf{3.22}  & 37.28 & 2.97  \\
B & Coreset      &  \xmark        & 40.18            & 3.12   & \textbf{ 39.63}  &  \textbf{3.10} \\
S & Random-SSL    &  \cmark        & 37.80            & 2.99  & 36.46  & 2.90 \\
S & Entropy-SSL   &   \cmark      & 38.32            & 3.03  &  36.70 & 2.93  \\
S & EqualAL-SSL   &  \cmark        & 39.43            & 3.07  & 36.31  & 3.06 \\
B & Coreset-SSL   &  \cmark        & 39.28            & 3.08  & 39.20  & 3.06 \\
\hline
- & 100\%         &  \xmark        & 56.87            & 4.55    & 48.85   & 3.91 \\
\hline
% \multicolumn{5}{|l|}{\textit{With MC-Dropout decoder} } \\
% \hline
% S & BALD        &   \xmark        &  \textbf{40.96} &  \textbf{3.19}      \\
% S & BALD-SSL     &  \cmark        &  39.27         &  3.15      \\
% %B & Coreset-MCD    & \xmark        &  40.36         &  3.14      \\
% %B & Coreset-MCD-SSL   &  \cmark     &  39.72         &  3.12      \\
% - & 100\%-MCD   &   \xmark        & 56.47          &  4.52    \\
% \hline 
\end{tabular}

\caption{Active Learning results on A2D2 Pool-0f in 2-2 setting. AUC@10 and mIoU@6 metrics are reported. A denotes Acquisition method type. S and B denotes the single-sample and batch-based acquisition, respectively.}
\label{table:a2d2_p0_2_2}
\end{table*}

\begin{table}[t]
\centering
\setlength\tabcolsep{3.5pt}
 \begin{tabular}{|c|@{\hspace{1.5mm}}l@{\hspace{1.5mm}}|@{\hspace{0.15cm}}c@{\hspace{0.15cm}}|@{\hspace{0.15cm}}c@{\hspace{0.15cm}}c@{\hspace{0.15cm}}|} 
 \hline 
 \multirow{2}{*}{\textbf{A}} & \textbf{AL Method}   & \multirow{2}{*}{\textbf{SSL}} & \multicolumn{2}{c@{\hspace{1.5mm}}|}{\textbf{Cityscapes: 2-2}}   \\  
  
%    \cmidrule(l{0.0em}r{4.0em}){2-3}  \cmidrule(r{4.0em}){4-5} \cmidrule(r{0.5em}){6-7}
    & \textbf{Metric} $\to$ &   & \textbf{mIoU@6}  & \textbf{AUC@10} \\
 [0.5ex] 
 \hline
S & Random      &   \xmark        & 46.05              & 3.65    \\ 
S & Entropy      &  \xmark        & \textbf{51.24}     & \textbf{4.00}    \\
B & Coreset       &  \xmark        & 47.26              & 3.74    \\
S & Random-SSL     &  \cmark       & 47.46              & 3.72    \\
S & Entropy-SSL     &  \cmark     & 49.99              & 3.93    \\
B & Coreset-SSL     &  \cmark     & 48.51              & 3.82    \\
\hline
- & 100\%       &   \xmark        & 68.42              & 5.47    \\ 
\hline
\end{tabular}

\caption{Active Learning results on Cityscapes dataset in low-budget 2-2 setting. AUC@10 and mIoU@6 metric are reported. A denotes Acquisition method type. S and B denotes the single-sample and batch-based acquisition, respectively.}
\label{table:city_2_2}
\end{table}

% \begin{figure*}[t!]
% \centering
% \begin{tabular}{r@{\hspace{1mm}}c}
% \begin{subfigure}{0.45\textwidth}
%  \includegraphics[trim={0 0.5cm 0 0cm}, clip, width=65mm]{./figures/a2d2/a2d2_p5.pdf}  \caption{A2D2:Pool-5f}
%  \end{subfigure} 
%  &
%   \begin{subfigure}{0.45\textwidth}
%  \includegraphics[trim={0 0.5cm 0 0cm}, clip, width=65mm]{./figures/a2d2/a2d2_p21.pdf} \caption{A2D2:Pool-21f}
%   \end{subfigure}  
% \end{tabular}
% \caption{Results on redundant datasets. Active Learning performance curves on A2D2 dataset: Pool-5f and Pool-21f. The X-axis shows the percentage of labeled datasets. The methods which utilize MC-Dropout in their network architecture are marked with $\ast$. Other performance curves for A2D2 Pool-11f and Pool-Aug are included in the main manuscript.}
% \label{fig:a2d2_pools_}
% \end{figure*}

% \begin{figure}[t]
% \centering
% \begin{tabular}{c@{\hspace{1mm}}c}
%  \includegraphics[trim={0 0cm 0 0cm}, clip, width=70mm]{figures/method_figure/Redundancy_wo_text.PNG}
% \end{tabular}
% \caption{A2D2 Pool-Aug dataset. Left: the original image. Right: the duplication through color augmentation and random cropping of the original image}
% \label{fig:Crop_and_augment}
% \end{figure}

\section{Results: AL under Different Budgets}
Here, we show the performance curves and tables for different active learning methods in a low annotation budget setting. This section also contains the remaining curves and tables in a high annotation budget setting, discussed in the main paper.

% \begin{itemize}
%     \item Add curves for 5-5 and 10-10 settings
% \end{itemize}

\subsection{Low Budget}

For diverse datasets, we evaluate active learning acquisition methods on PASCAL-VOC, A2D2-Pool-0f, and Cityscapes datasets. For redundant datasets, we show results on the A2D2-Pool-11f dataset only. 

\noindent\textbf{PASCAL-VOC:} We show results for the AL methods in a low-budget setting on the PASCAL-VOC dataset. We consider the 2-2 setting as the low-budget setting, where the maximum annotation budget is 10\% of the dataset size. We find that the random-SSL method performs the best. It shows that none of the AL bias is correctly learned or helpful in such a low-budget setting for such a diverse dataset. The integration of semi-supervised learning with random selection improves the performance over the supervised random sampling baseline. Results are shown in Table~\ref{table:pascal_2_2} and Figure~\ref{fig:pascal_2_2}.

\noindent\textbf{Cityscapes:}
Table~\ref{table:city_2_2} show results on the Cityscapes dataset in a low annotation budget, 2-2 setting. We find that the single-sample-based method performs the best. SSL integration with active learning is only useful for the batch-based Coreset approach, whereas it is detrimental for other acquisition functions.

\noindent \textbf{A2D2 Pool-0f:}
We show results in the low-budget, 2-2 setting for the A2D2 Pool-0f. We find that single-sample-based methods outperform all the methods. This shows that active learning is again successful in a low-budget setting when the dataset only covers a specific domain, like the driving scenario in this case. However, semi-supervised learning does not help in this case. Results are shown in Table~\ref{table:a2d2_p0_2_2} and Figure~\ref{fig:a2d2_2_2}.

\noindent\textbf{A2D2 Pool-11f:}
Table~\ref{table:a2d2_p0_2_2} show results on redundant datasets A2D2-pool-11f in low annotation budget, 2-2 setting. We find that batch-based methods outperform all the methods. 
Redundant datasets still favor the batch-based acquisition method in the low-budget setting. 
However, SSL does not contribute any additional improvements due to insufficient labeled samples to support learning from unlabeled samples.

As discussed in the main paper, active learning methods are highly sensitive to distribution change w.r.t. levels of redundancy in the low-budget setting. The ideal policy transitions from random selection to single-sample acquisition and then to batch-based acquisition as the level of redundancy in the dataset goes from low to high.

\subsection{High-budget}
\noindent\textbf{PASCAL-VOC:}
Figure~\ref{fig:pascal_10_10} and Figure~\ref{fig:pascal_5_5} show the AL performance curves for 10-10 and 5-5 settings on the PASCAL-VOC dataset, respectively. We observe that the single-sample-based methods with semi-supervised learning perform the best. Performance tables for these settings are included in the main paper.

% \noindent\textbf{A2D2:} Figure~\ref{fig:a2d2_pools_} shows the AL performance curves for 10-10 setting on A2D2-Pool5f and A2D2-Pool-21f datasets. For all redundant datasets, the Coreset-SSL approach consistently performs the best. Performance tables for these settings are included in the main paper.

\end{document}

\begin{tikzpicture}
  \centering
  \begin{axis}[
        xbar, 
        xbar=1.0pt,
        bar width=0.25cm,
        xmajorgrids, tick align=inside,
        major grid style={draw=black!20!white},
        enlarge x limits = {value = .1, upper},
        xmin=-3, xmax=3,
        axis y line*=middle,
        axis x line*=bottom,
        ylabel={Datasets},
        ylabel style={yshift=-0.5em,font=\scriptsize},
        xtick={-3,-2,-1, 0,1, 2,3},
        xticklabel style ={ font=\scriptsize},
        enlarge y limits = {abs=0.8},
        legend pos=north,
        legend style={nodes={scale=0.6, transform shape}, at={(0.9, 0.98)}, row sep=0.1cm, column sep=0.1cm, legend columns=1},
        legend image code/.code={
        \draw [#1] (0cm,-0.1cm) rectangle (0.4cm,0.08cm); },
        xlabel={Relative difference between Single-sample and Batch-based AL (\%)},
        xlabel style={yshift=0.4em, xshift=-2.0em, font=\scriptsize},
        %ytick={1,2,3,4,5,6,7,8},
        %yticklabels={PASCAL,Cityscapes,A2D2-Pool-0f,A2D2-Pool-5f, A2D2-Pool-11f,A2D2-Pool-21f,A2D2-Aug, A2D2},
        %symbolic y coords={pascal,city,a2d2_0f,a2d2_5f, a2d2_11f, a2d2_21f,a2d2_aug,a2d2},
        yticklabels = {PASCAL-VOC,Cityscapes,A2D2-Pool-0f, A2D2-Pool-5f, A2D2-Pool-11f,A2D2-Pool-21f,A2D2-Aug, A2D2},
        yticklabel style = {xshift=-82pt, yshift=0pt, font=\scriptsize},
        ytick={0.7,1.7,2.7,4.3,5.3,6.3,7.3,8.3},
        %ytick distance=1,
        %/pgf/bar shift={1pt},
    ]
    
    \addplot[draw=none, fill=blue!30] coordinates {
      (0.57,1)
      (2.50,2)
      (2.66,3)};
      \addplot[draw=blue!30, fill=blue!30, postaction={
pattern=north west lines, pattern color=black!60}] coordinates {
      (0.22,1)
      (-1.34,2)
      (1.96,3)};
    \addplot[draw=none, fill=red!40] coordinates {
      (-0.24,4)
      (-0.61,5)
      (-0.77,6)
      (-0.43,7)};
    \addplot[draw=red!40, fill=red!40, postaction={
pattern=north west lines, pattern color=black!60}] coordinates {
      (-0.95,4)
      (-2.98,5)
        (-1.08,6)
      (-2.18,7)};

%      \addplot[draw=none, fill=blue!40] coordinates {
%       (0.57,pascal)
%       (2.50,city)
%       (2.66,a2d2_0f)};
%       \addplot[draw=blue!40, fill=blue!40, postaction={
% pattern=north west lines, pattern color=black!70}] coordinates {
%       (0.22,pascal)
%       (-1.34,city)
%       (1.96,a2d2_0f)};
%     \addplot[draw=none, fill=red!40] coordinates {
%       (-0.24,a2d2_5f)
%       (-0.61,a2d2_11f)
%         (-0.77,a2d2_21f)
%       (-0.43,a2d2_aug)};
%     \addplot[draw=red!40, fill=red!40, postaction={
% pattern=north west lines, pattern color=black!70}] coordinates {
%       (-0.95,a2d2_5f)
%       (-2.98,a2d2_11f)
%         (-1.08,a2d2_21f)
%       (-2.18,a2d2_aug)};
       %(-1.44,a2d2)};
       
    \legend{Diverse dataset, ,Redundant dataset, }
    \node (G) at (480, 350) {\textcolor{black}{\scriptsize  Single-sample acquisition}};
     \node (G) at (480, 310) {\textcolor{black}{\scriptsize  Better}};
    \draw[>=latex, ->,black] (380,270) -- (580,270);
    \node (G) at (150, 230) {\textcolor{black}{\scriptsize  Batch-acquisition}};
    \node (G) at (150, 190) {\textcolor{black}{\scriptsize  Better}};
    \draw[>=latex, ->,black] (230,270) -- (50,270);
    \node [above right] at (rel axis cs:0.495,0.7) {\includegraphics[width=2.5cm]{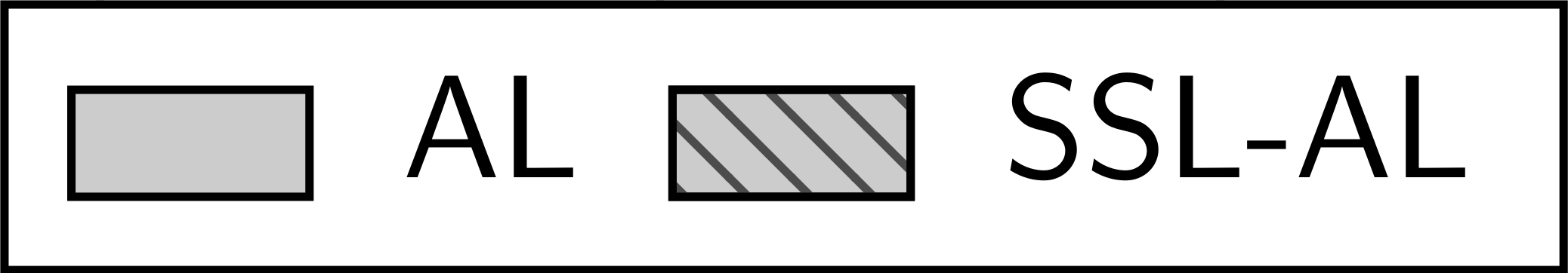}};
       % 1.46

    % \addplot[] [draw=none, fill=red!40] coordinates {
    %   (-0.96, a2d2_5f) % 0.96
    %   (-1.46, a2d2_11f) % 3.07
    %   (-3.0, a2d2_3k)}; % 1.46  
    %  \addplot[] [draw=none, fill=red!40] coordinates {
    %   (-0.96, 0) % 0.96
    %   (-1.46, 1) % 3.07
    %   (-3.0, 2)}; % 1.46  
      
    %   \node (A) at (-10, 560) {\textcolor{black}{\tiny  PASCAL}};
    %   \node (B) at (80, 785) {\textcolor{black}{\tiny  Cityscapes}};
    %   \node (C) at (200, 720) {\textcolor{black}{\tiny  A2D2:Pool-0f}};
    %   \node (D) at (320, 260) {\textcolor{black}{\tiny  A2D2:Pool-5f}};
    %   \node (E) at (410, 217) {\textcolor{black}{\tiny  A2D2-3k }};
    %   \node (F) at (510, 70) {\textcolor{black}{\tiny A2D2:Pool-11f }};
    %   \node (G) at (80, 290) {\textcolor{green!40!black}{\scriptsize  Diverse datasets}};
    %   \node (H) at (410, 510) {\textcolor{green!40!black}{\scriptsize  Redundant datasets}};
    %   \draw [decorate, decoration = {brace}] (200,360) -- (-40,360);
    %   \draw [decorate, decoration = {brace}] (300,430) -- (530,430);
    %   \draw[>=latex, <->,green!40!black] (150,18) -- (255,18);
    %   \node (I) at (320, 18) {\textcolor{green!40!black}{\tiny  Redundancy}};
    %   \node (J) at (100, 18) {\textcolor{green!40!black}{\tiny  Diversity}};
  \end{axis}
  \end{tikzpicture}